\documentclass[lettersize,journal]{IEEEtran}
\usepackage{amsmath,amsfonts}
\usepackage{algorithmic}
\usepackage{algorithm}
\usepackage{bigstrut}
\usepackage{subfigure}
\usepackage{array}
\usepackage{epsfig}
\usepackage[caption=false,font=normalsize,labelfont=sf,textfont=sf]{subfig}
\usepackage{textcomp}
\usepackage{stfloats}
\usepackage{url}
\usepackage{verbatim}
\usepackage{graphicx}
\usepackage{cite}
\usepackage{multicol}
\usepackage{multirow}
\def\BibTeX{{\rm B\kern-.05em{\sc i\kern-.025em b}\kern-.08em
    T\kern-.1667em\lower.7ex\hbox{E}\kern-.125emX}}

\title{SNN2ANN: A Fast and Memory-Efficient Training Framework for Spiking Neural Networks}
\author{Jianxiong Tang, Jianhuang Lai, \IEEEmembership{Senior Member, IEEE}, Xiaohua Xie, \IEEEmembership{Member, IEEE}, \\\quad \quad \quad Lingxiao Yang, and Wei-Shi Zheng, \IEEEmembership{Member, IEEE}
 \thanks{This work was supported by the Key-Area Research and Development Program of Guangzhou (202007030004), China.}
 \thanks{Corresponding author: Jianhuang Lai. }
 \thanks{Jianxiong Tang and Lingxiao Yang are with the School of Computer Science and Engineering, Sun Yat-sen University, Guangzhou 510006, China (e-mail: tangjx6@mail2.sysu.edu.cn; yanglx9@mail.sysu.edu.cn).}
 \thanks{Jianhuang Lai, Xiaohua Xie, and Wei-Shi Zheng are with the School of Computer Science and Engineering, Sun Yat-sen University, Guangzhou 510006, China, also with the Guangdong Province Key Laboratory of Information Security Technology, Sun Yat-sen University, Guangzhou, China, and also with the Key Laboratory of Machine Intelligence and Advanced Computing, Ministry of Education, China (e-mail: stsljh@mail.sysu.edu.cn; xiexiaoh6@mail.sysu.edu.cn; wszheng@ieee.org).}
}
\begin{document}

\maketitle

\begin{abstract}
Spiking neural networks are efficient computation models for low-power environments. Spike-based BP algorithms and ANN-to-SNN (ANN2SNN) conversions are successful techniques for SNN training. Nevertheless, the spike-base BP training is slow and requires large memory costs. Though ANN2NN provides a low-cost way to train SNNs, it requires many inference steps to mimic the well-trained ANN for good performance. In this paper, we propose a SNN-to-ANN (SNN2ANN) framework to train the SNN in a fast and memory-efficient way. The SNN2ANN consists of $2$ components: a) a weight sharing architecture between ANN and SNN and b) spiking mapping units. Firstly, the architecture trains the weight-sharing parameters on the ANN branch, resulting in fast training and low memory costs for SNN. Secondly, the spiking mapping units ensure that the activation values of the ANN are the spiking features. As a result, the classification error of the SNN can be optimized by training the ANN branch. Besides, we design an adaptive threshold adjustment (ATA) algorithm to address the noisy spike problem. Experiment results show that our SNN2ANN-based models perform well on the benchmark datasets (CIFAR10, CIFAR100, and Tiny-ImageNet). Moreover, the SNN2ANN can achieve comparable accuracy under $0.625\times$ time steps, $0.377\times$ training time, $0.27\times$ GPU memory costs, and $0.33\times$ spike activities of the Spike-based BP model. Code is available at \url{https://github.com/TJXTT/SNN2ANN}.
\end{abstract}

\section{Introduction}

\IEEEPARstart{I}n recent years, deep artificial neural networks \cite{lecun2015deep} (ANNs) have achieved outstanding performance for many applications such as image recognition \cite{DBLP:conf/nips/KrizhevskySH12,DBLP:conf/icml/YangZLX21}, biometric identification \cite{DBLP:conf/cvpr/SchroffKP15,guangcong2019aaai}, object detection \cite{DBLP:conf/cvpr/RedmonDGF16,DBLP:conf/iccv/LinGGHD17,DBLP:conf/aaai/KimPNY20}, etc. However, the success of deep ANNs relies on large-scale power consumption, which is not easy to deploy the ANNs on low-power platforms \cite{painkras2013spinnaker,DBLP:journals/micro/DaviesSLCCCDJIJ18,pei2019towards,9353400,9316906}. Spiking Neural Networks \cite{DBLP:journals/nn/TavanaeiGKMM19} (SNNs) is a promising learning model due to its computational efficiency for discrete spike events, which plays an important role in real-time practical applications \cite{10.3389/fnins.2017.00682}. The spiking neurons are the basic components of SNNs. These neurons fire the spikes only when the accumulated inputs exceed the threshold, making SNNs more power-efficient than ANNs. However, the outputs of the spiking neurons are binary and non-differentiable, disabling the traditional backpropagation training algorithms for SNNs.

Many algorithms have been proposed for training SNNs, and the most popular learning methodologies can be summarized in two classes: \textbf{a) Spike-based backpropagation algorithms} and \textbf{b) ANN-to-SNN conversion (ANN2SNN)}. The Spike-based backpropagation \cite{wu2019direct,DBLP:conf/cvpr/0006S020,DBLP:journals/neco/ZenkeV21} trains the SNN directly. One of the popular Spike-based BP algorithms is Spatio-temporal backpropagation
\cite{wu2019direct} (STBP). The STBP captures the Spatio-temporal dynamic of the SNNs, leading to few time steps for inference. However, the STBP requires the BP algorithm to propagate the gradients through spatial and temporal directions, resulting in a high computational cost for training.

To avoid the costly training of Spike-based BP algorithms, the ANN2SNN converts a trained ANN to the SNN version \cite{cao15,diehl2016conversion,10.3389/fnins.2019.00095}, which requires few parameters to train, and the conversion cost is low. Benefiting from the good performance of the well-trained ANN, ANN2SNN conversions perform good in many tasks. Nevertheless, the excellent performance of ANN2SNN relies on a large number of time steps to simulate the behaviors of the ANN, which is inefficient during model inference. In addition, the activation values and positions of the feature maps of SNNs are much different from the ANNs, resulting in the noisy spike problem and degenerating the performance of SNNs.

In this paper, we propose a SNN-to-ANN (SNN2ANN) framework to train the SNNs in a fast and memory-efficient way. The SNN2ANN consists of $2$ components: \textbf{a) a weight-sharing architecture between ANN and SNN}, and \textbf{b) spiking mapping units}. Firstly, the architecture enables the BP training on the ANN branch, and the weight-sharing mechanism guarantees that both SNN and ANN branches are updated simultaneously. Secondly, the spiking mapping units contain Rectified Spiking Unit (ReSU) and Straight-Through Spiking Unit (STSU), which adjust the spiking features as the activation values of the ANN branch. As a result, we can optimize the classification error of the SNN by training the ANN branch. Moreover, to address the noisy spike problem, we propose the adaptive threshold adjustment algorithm to decrease the firing number of noisy spikes. Our SNN2ANN differs from the ANN2SNN conversions in two aspects: a) the SNN2ANN does not require the spiking neurons to simulate the behaviors of ANN; b) the spiking mapping units bound the time steps to result in fast inference. In addition, the spatial-temporal gradient propagation of the Spike-based BP is not involved in our SNN2ANN, reducing the GPU memory consumption and improving the training speed.

Our main contributions can be summarized below:
\begin{itemize}
    \item We design a SNN2ANN framework to transfer the training of SNN on the ANN. Different from the Spike-based BP that directly trains the SNN, the SNN2ANN trains the SNN on the ANN branch, reducing the training time and memory costs.
    \item We propose the Rectified Spiking Unit (ReSU) and Straight-Through Spiking Unit (STSU) to model the ANN branch. The ReSU/STSU establishes the equivalence relation between the ANN and SNN branches. When the ANN training is finished, we can obtain the SNN with a similar performance to the ANN branch.
    \item We propose the adaptive threshold adjustment (ATA) algorithm to address the noisy spike problem. The ATA adaptive increases the firing threshold of the IF neuron based on the noisy activation during training. The increased threshold decreases the firing number of noise spikes, which improves the performance of SNNs.
\end{itemize}

Experiment results demonstrate that our SNN2ANN performs well on CIFAR10, CIFAR100, and Tiny-ImageNet datasets with a few time steps and significantly saves training consumption. For example, the SNN2ANN can achieve comparable accuracy under $0.625\times$ time steps, $0.377\times$ training time, $0.27\times$ GPU memory costs, and $0.33\times$ spike activities of the Spike-based BP model.
\section{Background\&Related Work}
\subsection{Spiking Neural Networks}
 The Integrate-and-Fire (IF) neuron is the basic component of SNN. Given a fully connected SNN with $L$ layers, the dynamic of the $i$-th neuron of layer $n$ can be described as
\begin{align}\label{eq:if}
&u_i^{t,n} = u_i^{t-1,n}(1-o_i^{t-1,n}) + \sum\limits_{j=1}^{l_{n-1}} w_{j,i}^{n}o_{j}^{t,n-1}+b_i^n,\\
&o_i^{t,n} =\left\{\begin{array}{l}
1, \text { if } {u}_{i}^{t,n} > V_{th},  \\
0, \text { otherwise },
\end{array}\right.
\end{align}
where $o_i^{t,n}$ denotes the spike and $u_i^{t,n}$ is the membrane potential (MP) at $t$. $w_{j,i}$ is the weight connects the $j$-th neuron in the layer $n-1$ and the $i$-th neuron in layer $n$, and $b_i^n$ is the bias. The MP integrates the pre-synaptic spikes in the temporal direction, and the post-synaptic spikes are generated when the MP crosses $V_{th}$. Such a Spatio-temporal dynamic makes SNNs much different from the popular ANNs (e.g., CNN and LSTM). In addition, the discreteness of the spiking output disables the gradient calculation for BP training.

\begin{figure}[t]
    \centering
    \includegraphics[width=0.45\textwidth]{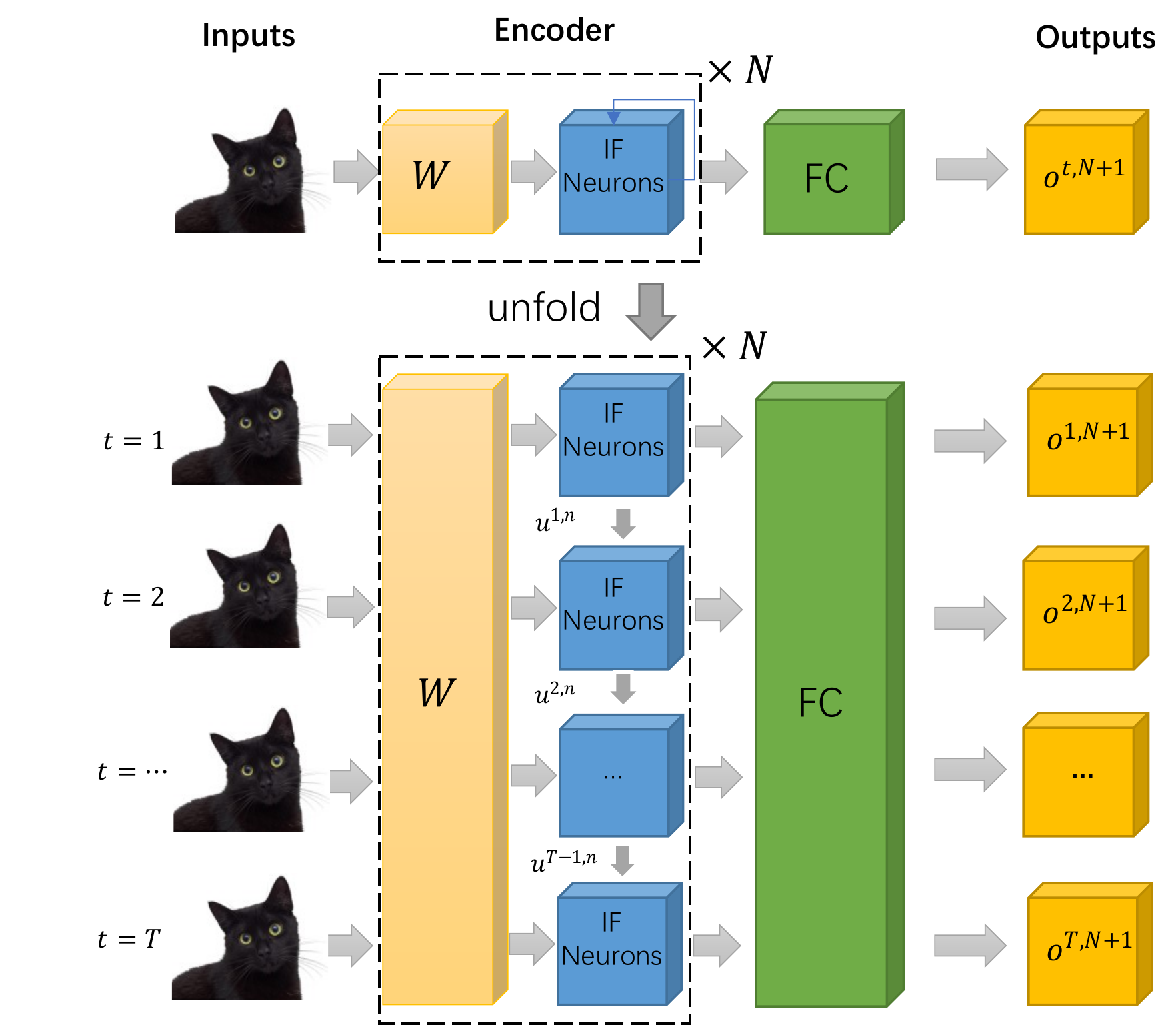}
    \caption{The Spatio-temporal dynamics of the SNN with IF neurons. We simulate the input stimulus by repeat the input image $T$ times.}
    \label{fig:if}
\end{figure}

Fig.~\ref{fig:if} shows the forward propagation of the SNN with IF neurons in $T$ steps. The input layer receives a sequence of images for the encoder layers. For an encoder layer, the synaptic weights integrate the outputs from the previous encoder, and then the IF neurons generate the spikes to the next layer and accumulate the MP for the next time step. Finally, the output layer integrates the outputs of the last encoder and makes a decision for classification.
\subsection{Spike-based Backpropagation}
The obstacle for SNN training with the BP algorithm is that the gradient of spike w.r.t the MP is $0$ everywhere. Therefore, many surrogate gradients \cite{2018Spatio,jin2018hybrid,lee2019enabling,Fang_2021_ICCV,DBLP:journals/corr/abs-2105-08810,DBLP:conf/icml/YangZL21,DBLP:journals/neco/ZenkeV21} that enable the spike gradient calculation are designed to enable the BP training for SNNs. One of the popular Spike-based BP algorithms is Spatial-Temporal backpropagation \cite{2018Spatio,wu2019direct} (STBP), which designs a rectangle function as the spike gradient for backward propagation.
\begin{equation}\label{eq:stbp}
    \frac{\partial o^{t,n}}{\partial u^{t,n}} = \frac{1}{a}\text{sign}(|u^{t,n}-V_{th}|<\frac{a}{2}),
\end{equation}
where $a>0$ is a scalar factor. The Spike-based BP training captures the spatial-temporal dynamic of the SNN, resulting in small time steps for inference. However, based on Eq.~\eqref{eq:if}, the gradients propagate through the spatial and temporal directions for synaptic weights updating. By detaching the gradient calculation of $o_i^{t-1,n}$ in Eq.~\eqref{eq:if}, for $1<t\leq T$, we can write the gradient of $o_i^{t,n}$ w.r.t $w_{j,i}^n$ as
\begin{align}\label{eq:spike_based_bp}
    \frac{\partial o_i^{t,n}}{\partial w_{j,i}^n}& = \frac{\partial o_i^{t,n}}{\partial u_{i}^{t,n}}\frac{\partial u_{i}^{t,n}}{\partial w_{j,i}^n}\nonumber\\
    &=  \frac{\partial o_i^{t,n}}{\partial u_{i}^{t,n}}(\sum\limits_{j=1}^{l_{n-1}}o_j^{t,n-1}+ (1-o_i^{t-1,n})\frac{\partial u_{i}^{t-1,n}}{\partial w_{j,i}^n}).
\end{align}
The gradients for $w_{j,i}^n$ are calculated in every time step, leading to a slow training speed and large memory cost.
\subsection{ANN-to-SNN Conversion}
The ANN2SNN conversion is to convert a trained ANN to its SNN version, which requires few parameters to train, and the conversion cost is low. Many conversion-based SNNs are built based on the spiking neuron with a soft reset \cite{DBLP:journals/corr/abs-2008-03658, DBLP:conf/ijcai/DingY0H21,DBLP:conf/aaai/YanZW21}, and the MP dynamic is
\begin{equation}
u_i^{t,n} = u_i^{t-1,n} + \sum\limits_{j=1}^{l_{n-1}} w_{j,i}^{n}o_{j}^{t,n-1}+b_i^n - o_i^{t-1,n}V_{th}.
\end{equation}
Then, the firing rate of the $i$th neuron of layer $n$ at step $T$ is
\begin{equation}\label{eq:snn_fr}
    r_i^{T,n} = \frac{\sum_{t=1}^To_i^{t,n}}{T} = \frac{\sum_{j=1}^{l_{n-1}} w_{j,i}^{n}r_j^{T-1,n}+b_i^n}{V_{th}}-\frac{u_i^{T,n}}{T V_{th}}.
\end{equation}
Based on the Rate coding mechanism and the soft reset, the ANN2SNN conversions treat the activation of ANN as the spiking firing rate:
\begin{equation}\label{eq:ann_fr}
    r_i^n = \min(\max(\frac{\sum_{j=1}^{l_{n-1}} w_{j,i}^{n} r_j^{n-1}+b_i^n}{V_{th}},0),1).
\end{equation}
Since $r_i^n\in [0,1]$, the SNN simulates the ANN by approximating the spiking firing rate layer-by-layer. However, we can find that the equivalent between Eq.~\eqref{eq:snn_fr} and \eqref{eq:ann_fr} is only established when $T\rightarrow+\infty$.
The conversion-based SNN requires many inference steps to mimic a well-trained ANN for good performance. Till now, many techniques have been proposed to shrink the converted SNN inference time steps, such as Spike norm \cite{10.3389/fnins.2019.00095}, Max norm \cite{rueckauer2017conversion}, Robust norm \cite{rueckauer2017conversion}, Rate norm \cite{DBLP:conf/ijcai/DingY0H21}, et al. Meanwhile, some coding mechanisms are designed to replace the rate coding of ANN2SNN for faster inference, such as Temporal-Switch Coding \cite{DBLP:conf/eccv/HanR20}, and FS-conversion coding \cite{naturearticle} and so on. However, the inference steps of ANN2SNN are much larger than the spike-based BP models.

Recently, \cite{9492305} proposed tandem learning to reduce the training cost of SNNs. Nevertheless, the noisy spikes of the SNN are not addressed, and the shallow structure limits the performance. The inference steps of the conversion-based SNN are large, while the spike-based BP training is slow and costs many GPU memories. It promotes us to design a fast and memory-efficient technique to train the deep SNNs.
\section{Methodology}
In this section, we design the SNN2ANN for SNN learning. We first analyze the noisy spike problem in the weight-shared SNN in Section III.A. Then, we propose the ANN2SNN framework to transfer the SNN training on ANN in Section III.B. In Section III.C, we design the spiking mapping units (ReSU and STSU) to rectify the spiking features on the ANN branch for training. Furthermore, the adaptive threshold adjustment is designed to address the noisy spike problem.

\subsection{Noisy Spikes in Weight-shared SNN}
 Transferring the trained weights of ANN on the SNN is a direct way for SNN learning. If the accumulated spikes of SNN are equivalent to the activation values of the ANN, the models should have the same performance. However, the inputs/outputs of the spiking neurons are binary sequences, leading to the activation values of SNN being much different from the ANN. The SNN may activate the neurons which are inactive in ANN. Such inconsistent activation neurons generate noisy spikes, resulting in the degeneration performance of the SNN.
 \begin{figure}[t]
    \centering
    \includegraphics[width=0.5\textwidth]{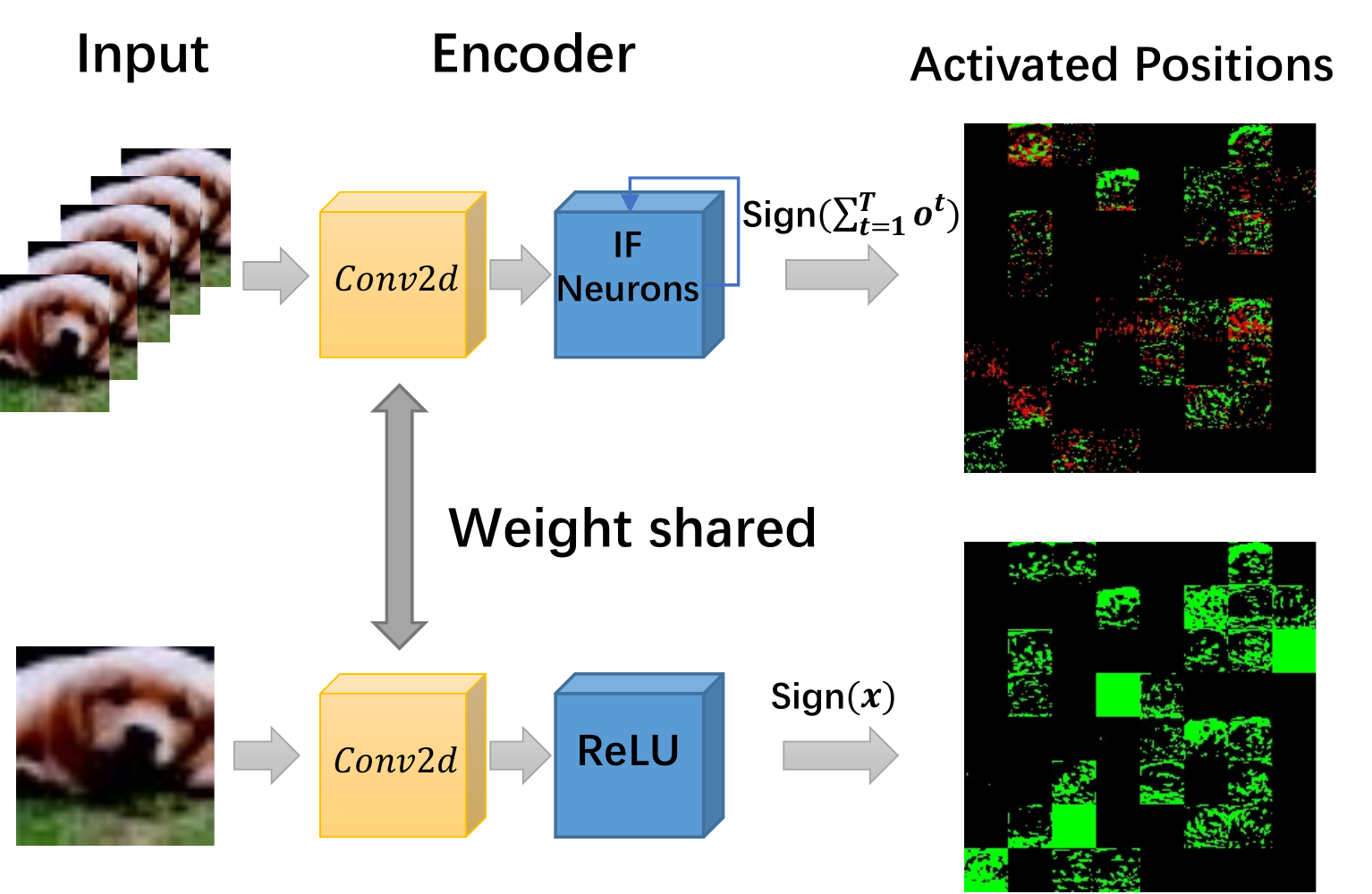}
    \caption{The activated positions of the weight sharing ANN and SNN. In the output of IF neurons, the green colors denote the same activated positions with ANN, and the red colors are the incorrect activates.}
    \label{fig:noisy}
\end{figure}
 \begin{figure}[t]
    \centering
    \includegraphics[width=0.52\textwidth]{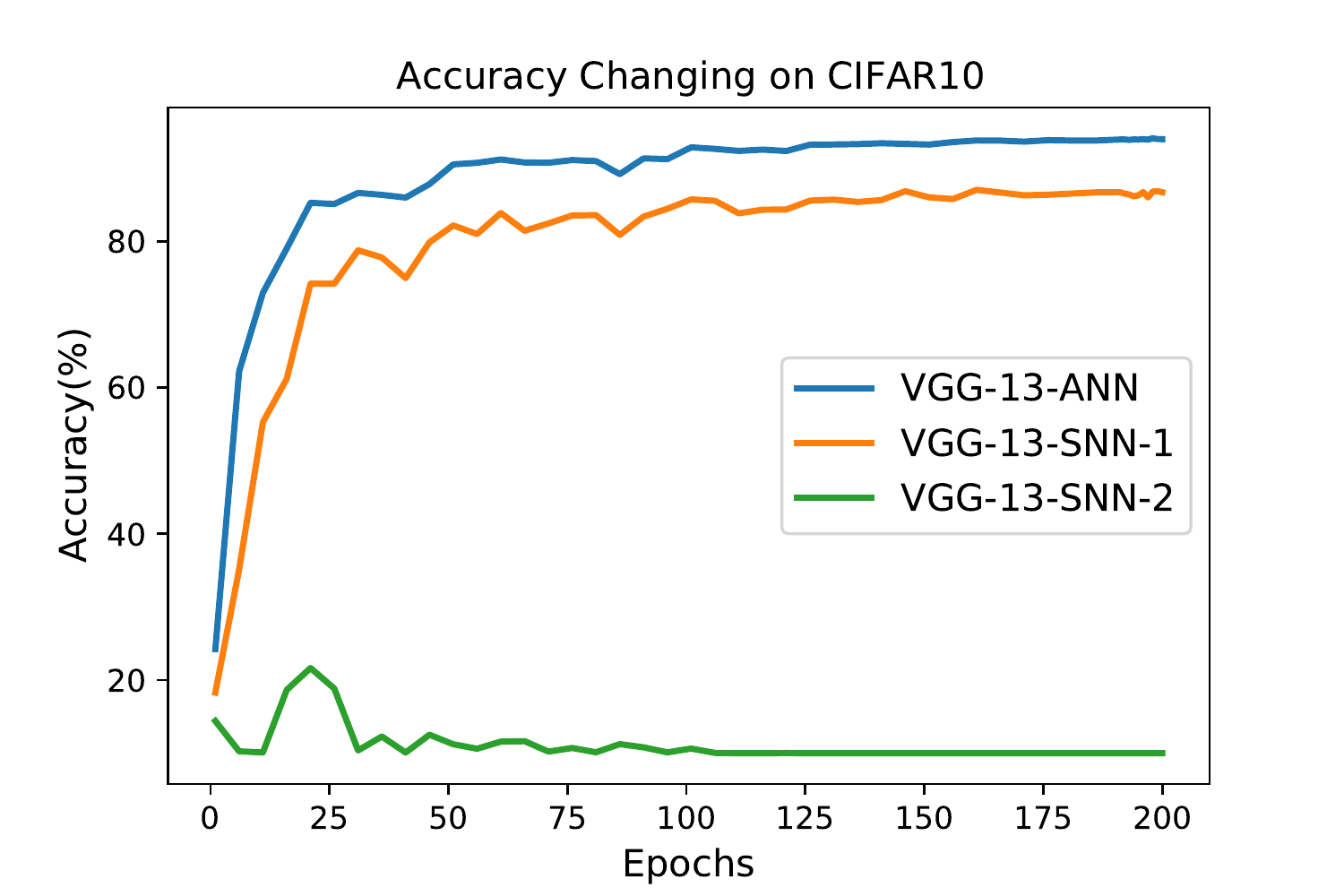}
    \caption{Accuracy comparison between the weight-shared ANN and SNNs. VGG-13-ANN is a CNN model. For the SNN models, the VGG-13-SNN-1 is denoised by the activated positions of the VGG-13-ANN, and VGG-13-SNN-2 is the SNN without denoising.}
    \label{fig:fig3}
\end{figure}

 Fig.~\ref{fig:noisy} visualizes the activated positions of the weight-shared SNN and ANN layer. The upper branch is the SNN, and the lower branch denotes the ANN. The $conv2d$ is a convolution layer, and the sign function is applied to obtain the activation positions from the outputs of the ReLU\cite{DBLP:conf/icml/NairH10}/IF neuron. Both SNN and ANN share the same input image. For the SNN, the input image is copied $5$ times as the sequence feed into the network. The integrate-and-fire mechanism makes the dynamic of the IF neuron much different from ReLU. The inactive positions in ReLU may be activated in the IF neuron, as the red points are shown in the up-right part of Fig.~\ref{fig:noisy}. These inconsistent activations in SNN are the noisy spikes, making the performance of the weight-shared SNN not comparable to ANN.

 We consider the weight-shared SNN and ReLU-based ANN using a VGG-13 backbone for a CIFAR10 training task and model the SNN with Integrate-and-Fire (IF) neurons. To demonstrate that the noisy spikes affect the performance of SNNs, we apply the ReLU activation positions of the ANN for the SNN to filter the inconsistent activation, and such SNN is denoted as ``VGG-13-SNN-1''. The SNN that is only weight-shared with the ANN is denoted as ``VGG-13-SNN-2''. Fig.~\ref{fig:fig3} shows an accuracy changing curve of the weight-shared ANN and SNNs. By utilizing the position information of the ReLU outputs, the VGG13-SNN-1 filters the noisy spikes and shows a similar changing tendency to the VGG-13-ANN. However, the VGG13-SNN-2 w/o denoising obtains poor performance. Filtering the noisy spikes is the key to improving the performance of SNN. However, it is hard to obtain the position information of ANN for SNN inference.
\subsection{SNN-to-ANN Framework}
\begin{figure}
    \centering
    \includegraphics[width=0.51\textwidth]{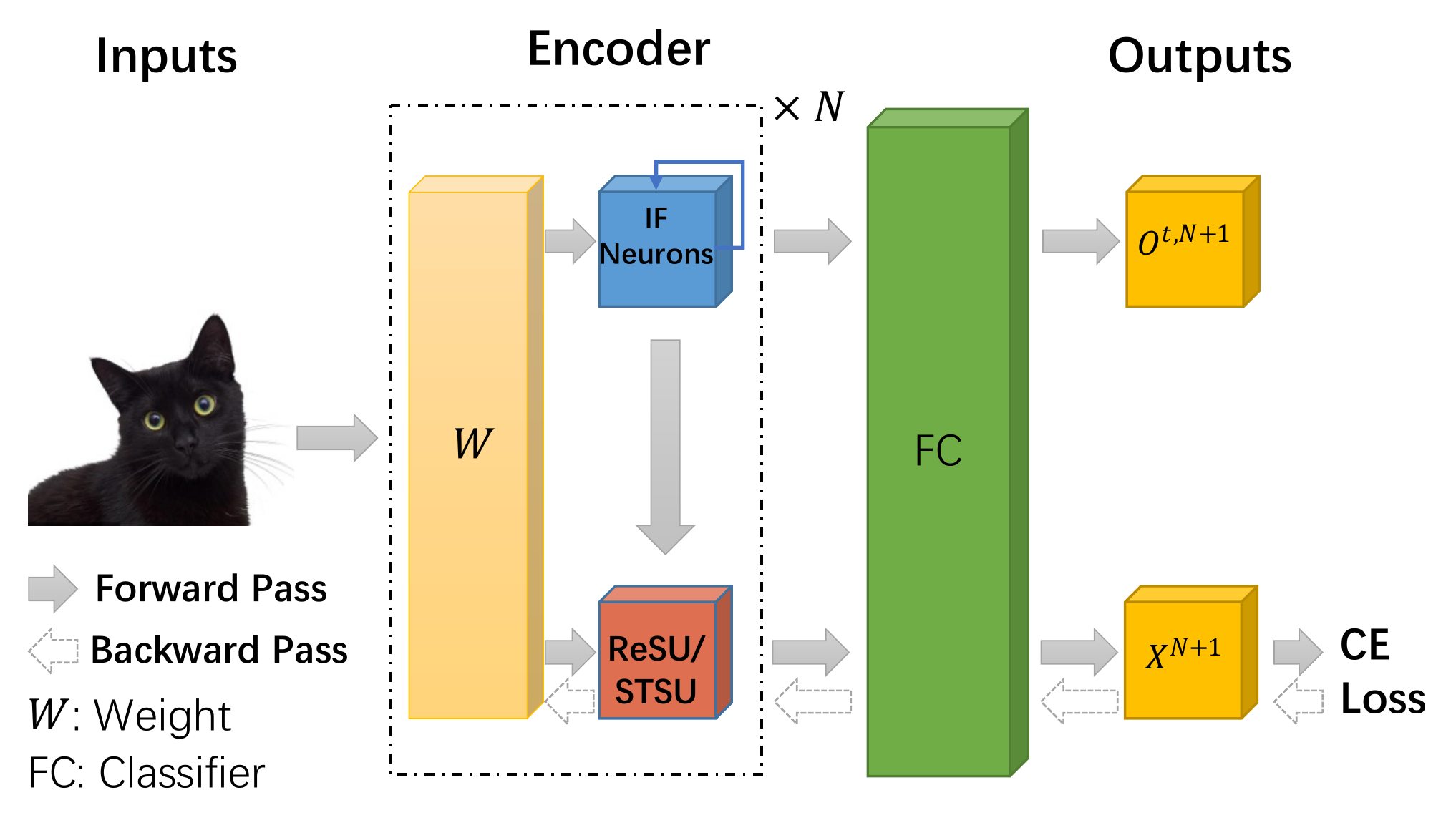}
    \caption{Overview of the SNN2ANN framework. The upper branch is the SNN with Integrate-and-Fire neurons, while the lower branch is the ReSU/STSU-based ANN. The ReSU/STSU incorporates the accumulated spikes on the ANN branch, and BP training works on the ANN branch for parameter updating. Both ANN and SNN branches share the weights and classifier.}
    \label{fig:fig1}
\end{figure}
The ANN2SNN conversions are not inference efficient, while the spike-based BP training requires too many computation resources. To address these problems, we design an SNN-to-ANN (SNN2ANN) framework to transfer the training of SNN on the ANN. The overview of the SNN2ANN is shown in Fig.~\ref{fig:fig1}. The upper branch is the SNN with IF neurons, while the lower branch is the ReSU/STSU-based ANN. $W$ denotes the sharing weight for ANN and SNN, and FC is the fully connected layer for classification. In the Encoder module, the ReSU/STSU maps the accumulated spikes on the ANN branch so that the features of SNN are incorporated into the training of ANN. Then, the BP training is worked on the ANN branch, and the weight sharing mechanism updates the parameters of the SNN branch. Compared with the spike-based BP training, the gradients in the SNN2ANN framework only propagate on the ANN branch, reducing the computation resource and training time. Since the activation values of ReSU/STSU are the accumulated spikes of the SNN, the max activation value is equivalent to the time steps. By setting the time steps to a small value, the SNN2ANN can learn a time-efficient SNN. We detail the principle of the SNN2ANN in the following sections.
\subsection{Spiking Mapping Units}
As mentioned in Section III.A, the activation positions of ANN can not be used for the weight-shared SNN. To address this problem, we share the spiking activation for the training of ANN. Our idea is to design the Rectified Spiking Unit (ReSU) and Straight-Through Spiking Unit (STSU) to transfer the spiking features of SNN for ANN branch modeling.
\subsubsection{Rectified Spiking Unit}
\begin{figure}[t]
    \centering
    \scalebox{1.0}{
    \includegraphics[width=0.45\textwidth]{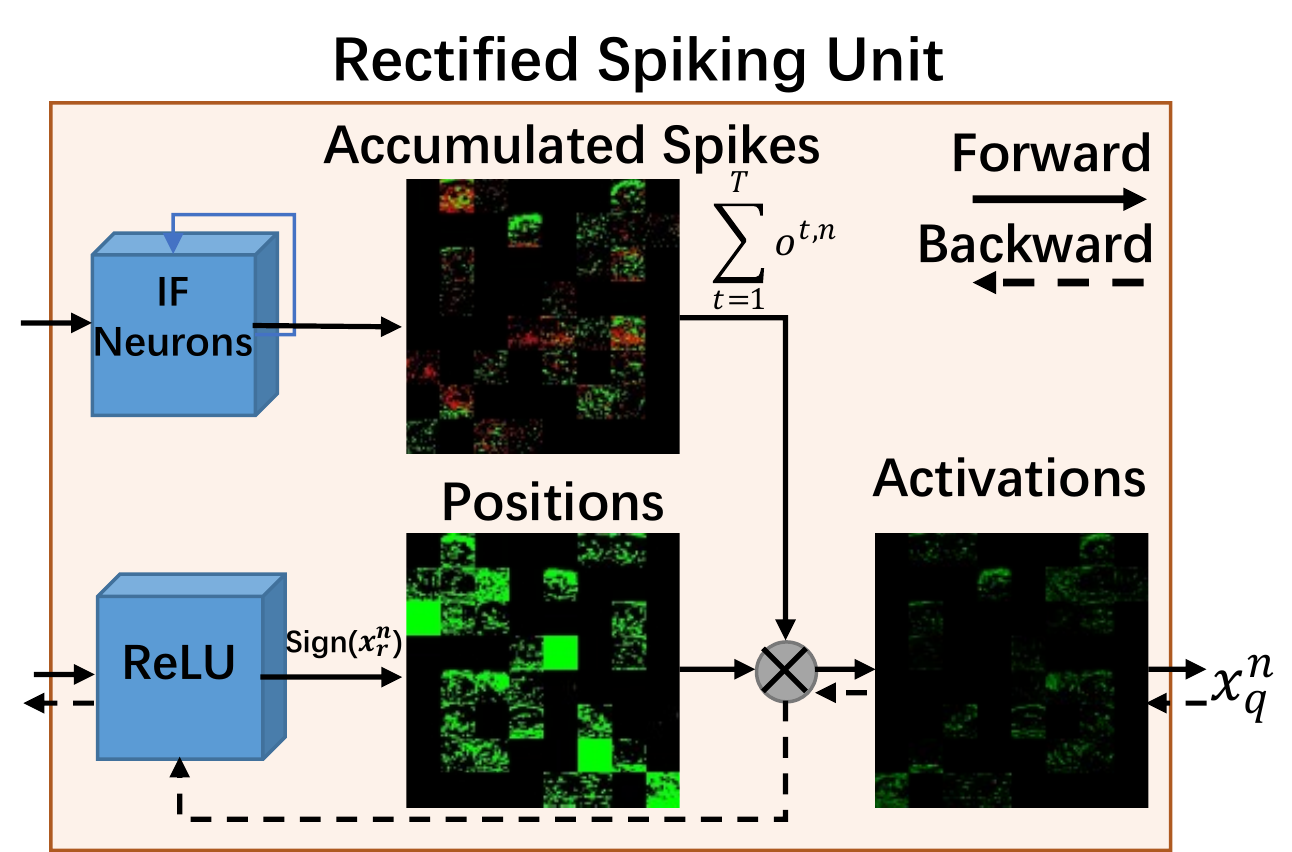}}
    \caption{The workflow of the Rectified Spiking Unit (ReSU).}
    \label{fig:resu}
\end{figure}

Let $x_q^n$ denotes the output of the $n$-th layer of the ANN branch, and the ReSU can be described as
\begin{equation}\label{eq:resu}
   x_q^n = \text{Sign}(x_r^n)\circ \sum\limits_{t=1}^T o^{t,n},
\end{equation}
where $x_r^n$ is the output of ReLU activation, $o^{t,n}$ is the spiking outputs of the $n$-th layer of the SNN at time step $t$, and $\circ$ denotes the Hardamand product. The \text{Sign}$(\cdot)$ maps the outputs of ANN in $\{0,1\}$ which can be regarded as the activated positions of the ReLU. Such position information disables the noisy spikes to participate in the training of ANN. With the ReSU, the filtered spiking features are propagated through the ANN branch so that the training error is calculated based on the spiking features. The workflow of the ReSU is shown in Fig.~\ref{fig:resu}.

Now, we give the gradient analysis on the ReSU activation. Suppose $\mathcal{L}$ is the loss function of the network and $W^n$ are the weights of layer $n$, then the gradient of $\mathcal{L}$ w.r.t $W^n$ is
\begin{align}
\centering
    \frac{\partial \mathcal{L}}{\partial W^n} =& \frac{\partial \mathcal{L}}{\partial x_q^n}\frac{\partial x_q^n}{\partial W^n}  \\ \nonumber
    =&\frac{\partial \mathcal{L}}{\partial x_q^n}\frac{\partial x_q^n}{\partial x_r^n}\frac{\partial x_r^n}{\partial W^n}+\sum\limits_{t=1}^T\frac{\partial \mathcal{L}}{\partial x_q^n}\frac{\partial x_q^n}{\partial o^{t,n}}\frac{\partial o^{t,n}}{\partial W^n}.
\end{align}
Since the training is worked on the ANN branch, the accumulated spikes are not involved in the gradient calculations for weight updating,
$$
\frac{\partial o^{t,n}}{\partial W^n}=0,~t\in[1,T].
$$
The Sign$(\cdot)$ quantifies the ReLU outputs to the position information, disabling the gradient propagation from  $\mathcal{L}$ to $W^n$. Therefore, we introduce the Straight-Through-Estimator \cite{bengio2013estimating} (STE) to address this problem. Specifically, the STE passes the gradient of $\mathcal{L}$ w.r.t $x_r^n$ to $x_q^r$,
\begin{equation}
    \frac{\partial \mathcal{L}}{\partial x_r^n} = \frac{\partial \mathcal{L}}{\partial x_q^n}.
\end{equation}
Then, the gradient for weights and biases updating becomes
\begin{align}
    &\frac{\partial \mathcal{L}}{\partial W^n} = \frac{\partial \mathcal{L}}{\partial x_q^n}\frac{\partial x_r^n}{\partial W^n}\label{eq:ste1},\\    &\frac{\partial \mathcal{L}}{\partial b^n} = \frac{\partial \mathcal{L}}{\partial x_q^n}\frac{\partial x_r^n}{\partial b^n}\label{eq:ste2}.
\end{align}
\subsubsection{Straight-Through Spiking Unit}
\begin{figure}[t]
    \centering
    \scalebox{1.0}{
    \includegraphics[width=0.45\textwidth]{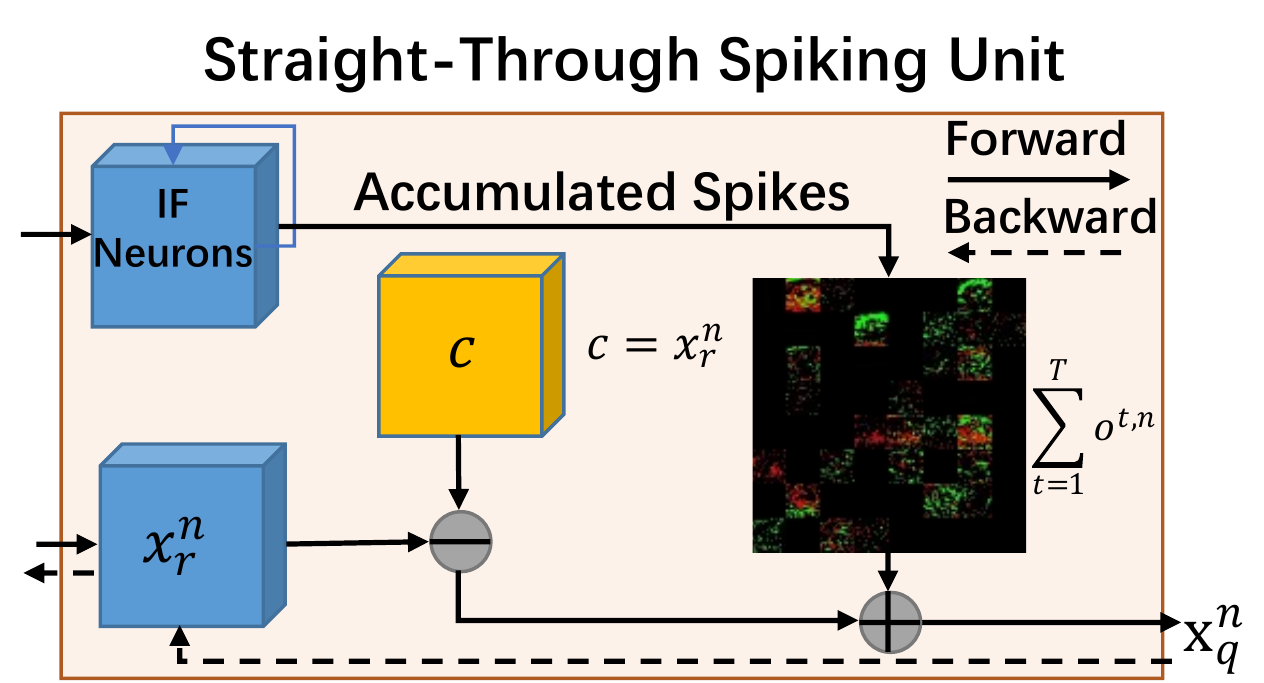}}
    \caption{The workflow of the Straight-Through Spiking Unit (STSU). $c$ is a constant that is equivalent to $x_r^n$.}
    \label{fig:stsu}
\end{figure}

Unlike the ReSU filters noisy spikes, the Straight-Through Spiking Unit (STSU) regards the activation positions of SNN are also activated in ANN and maps all spikes for the training of the ANN branch. The definition of STSU is
\begin{equation}\label{eq:masu}
    x_q^n = \sum\limits_{t=1}^T o^{t,n} + x_r^n - c,~s.t.~ c=x_r^n.
\end{equation}
The STSU maps the accumulated spikes $\sum_{t=1}^To^{t,n}$ on the ANN branch, and the true output $x_r^n$ of the ANN is offset by the constant $c$. Therefore, the STSU ensures that the outputs of the ANN branch are equivalent to $\sum_{t=1}^To^{t,n}$, and $x_r^n$ enables gradient calculation for BP training. The workflow of the STSU is presented in Fig. \ref{fig:stsu}, and the gradients of $\mathcal{L}$ w.r.t $W^n$ and $b^n$ are calculated based on STE:
\begin{align}
    &\frac{\partial \mathcal{L}}{\partial W^n} = \frac{\partial \mathcal{L}}{\partial x_q^n}\frac{\partial x_q^n}{\partial x_r^n}\frac{\partial x_r^n}{\partial W^n} = \frac{\partial \mathcal{L}}{\partial x_q^n}\frac{\partial x_r^n}{\partial W^n}\label{eq:stsu1},\\    &\frac{\partial \mathcal{L}}{\partial b^n} =  \frac{\partial \mathcal{L}}{\partial x_q^n}\frac{\partial x_q^n}{\partial x_r^n}\frac{\partial x_r^n}{\partial b^n} = \frac{\partial \mathcal{L}}{\partial x_q^n}\frac{\partial x_r^n}{\partial b^n}\label{eq:stsu2}.
\end{align}
Using ReSU/STSU for SNN2ANN modeling, the accumulated spikes propagate in the forward pass while the STE enables the BP training on ANN. In addition, the activation values of ReSU/STSU are bounded by the size of the time window, promoting us to design a time-efficient SNN. Compared with the Spike-based BP (Eq.~\eqref{eq:spike_based_bp}) that propagates the gradients through spatial and temporal direction, the SNN2ANN trains the SNN on the ANN branch, reducing the training time and memory cost. We anticipate that SNN2ANN can increase the training efficiency of the SNN.
\subsection{Adaptive Threshold Adjustment}
The ReSU/STSU provides the pseudo features for ANN training, but the noisy spikes remain on the SNN branch. Therefore, we design the Adaptive Threshold Adjustment (ATA) to address such a noisy spike problem during training:
\begin{eqnarray}\label{eq:ata}
    &V_{th}^{k+1} = V_{th}^k + \xi(1-\alpha)V_{th}^{k},\\
    &\xi = \tau\max(0, \text{Sign}(\frac{1}{|\Omega|}\sum\limits_{n\in \Omega}\sum\limits_{t=1}^T o_n^t-\varepsilon))),
\end{eqnarray}
where $k$ is the $k$-th iteration in a training epoch, $V_{th}^k$ is the threshold potential of the IF neuron, $\alpha \in [0,1]$, $\varepsilon \in [0,1]$ and $\tau>0$ are the is the momentum, tolerance and scalar factors, respectively. $\Omega$ is a set of the positions of noisy spikes. If the average number of the accumulated noises is larger than $\varepsilon$, Eq.~\eqref{eq:ata} increases $V_{th}$. The ATA adaptive mechanism increases the firing threshold so that the number of noise spikes of a neuron can be reduced.
 \begin{figure}[t]

    \includegraphics[width=0.48\textwidth]{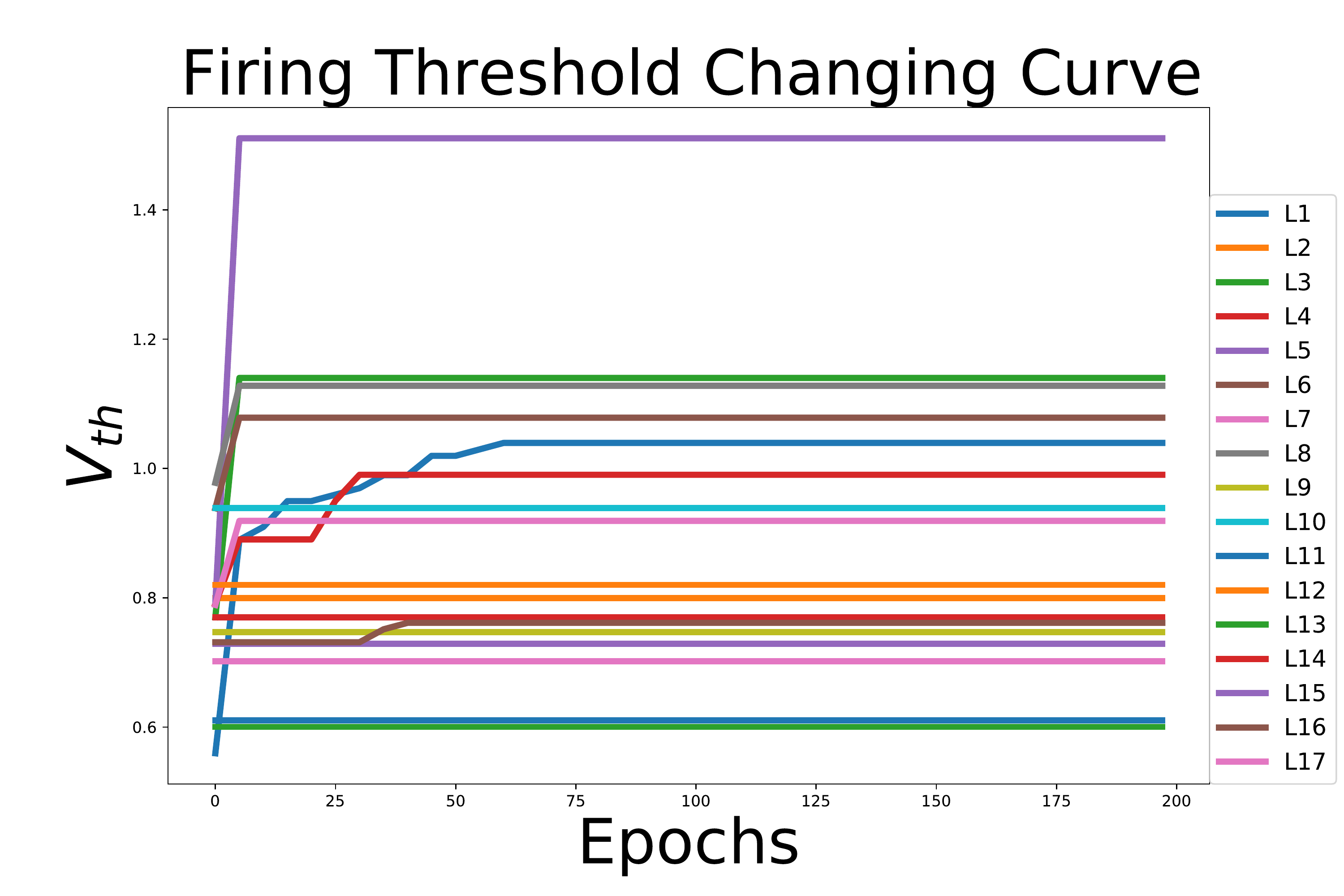}
    \caption{Layer-wise firing threshold changing of ResNet-17 with Adaptive Threshold Adjustment. ``$\text{L}_{n}$'' denotes the threshold of layer $n$. }
    \label{fig:ata_vth}
\end{figure}
We present the changing average firing threshold of each layer of the STSU-based ResNet-17 with $5$ time steps on CIFAR100 in Fig.~\ref{fig:ata_vth}. Each threshold is initialized to be a uniform distribution. It can be seen that all thresholds show a non-descent tendency. Based on Eq.~\eqref{eq:ata}, if the average of the noisy spikes is larger than $\varepsilon$, the firing threshold will increase. Otherwise, the threshold remains unchanged.
\subsection{Weight Updating with Batch Normalization}
Batch Normalization \cite{ioffe2015batch} (BN) is a technique to remove the internal covariate shift in deep networks training. In the training of ANN, the BN operation is
\begin{equation}
    \text{BN}(x_q^n)=\gamma\frac{x_q^n-\mu}{\sqrt{\sigma^2+\epsilon}}+\beta,
\end{equation}
where $\gamma$ and $\beta$ are the learnable parameters, and $\epsilon>0$ is a small enough number. $\mu$ and $\sigma$ are the mean and variance.  Due to the temporal structure of the SNN, the BN operation of ANN is not suitable for SNN. Therefore, we transfer the BN to the weights and biases with Exponential Moving Average:
\begin{align}
    &\text{BN}(W) = \gamma \frac{W}{\sqrt{\tilde{\sigma}^2+\epsilon}}\label{eq:bn1}, \\
    &\text{BN}(b) = \frac{b-\tilde{\mu}}{T\sqrt{\tilde{\sigma}^2+\epsilon}}+\frac{\beta}{T}\label{eq:bn2},
\end{align}
where $$\tilde{\mu} = (1-\alpha)\mu^{k-1}+\alpha\mu^k,\quad\tilde{\sigma} = (1-\alpha)\sigma^{k-1}+\alpha\sigma^k,$$ $T$ is time steps, and $k$ denotes the training iteration number.

Details of the SNN2ANN training pipeline are described in Algo.\eqref{algo:1}.
\begin{algorithm}[t]
\caption{SNN2ANN Training Pipeline.} %
{\bf Require:}  %
ANN: $f_a(\cdot)$, SNN: $f_s(\cdot)$, Training Set: $\{x_i,y_i\}_{i=1}^N$, Model parameters: $\{W^n,b^n, V_{th}^n\}_{n=1}^L$, ATA parameters: $\alpha\in[0,1]$, $\tau>0$ and $\varepsilon>0$, Time Steps: $T$, Epoch number: $E$, Learning rate: $\beta$, Loss function: $\mathcal{L}$.\\
{\bf Ensure:} $\{x_{q,i}^0\}_{i=1}^N = \{o_i^0\}_{i=1}^N=\{x_i\}_{i=1}^N, n\in [2,L]$.\\
{\bf Init:} $V_{th}^n\sim\mathcal{U}(0,1), n\in [1,L]$.
\begin{algorithmic}
\STATE \textbf{For} $e=1$ to $E$
\STATE \hspace{0.5cm} \textbf{Obtain the accumulated spikes from SNN branch.}
\STATE \hspace{0.5cm} \textbf{For} $t=1$ to $T$ %
\STATE \hspace{1.0cm} {\textbf{For} $n=1$ to $L$}
\STATE \hspace{1.5cm} \textbf{If} $t=1$
\STATE \hspace{2.0cm} $o^n=0$;
\STATE \hspace{1.5cm} Normalize $\{W^n, b^n\}$ based on Eq.~\eqref{eq:bn1}$\sim$ \eqref{eq:bn2};
\STATE \hspace{1.5cm} $W^n_s = \text{BN}(W^n), b^n_s = \text{BN}(b^n)$;
\STATE \hspace{1.5cm} Calculate $o^{t,n}$ with $\{W^n_s, b^n_s\}$ based on Eq.~\eqref{eq:if};
\STATE \hspace{1.5cm} Collect the spiking outputs: $o^n = o^n + o^{t,n}$;
\STATE \hspace{1.5cm} \textbf{If} $t=T$
\STATE \hspace{2.0cm} \textbf{Map the accumulated spikes on ANN branch.}
\STATE \hspace{2.0cm} Calculate the output of the $n$ th layer of ANN branch: $x_r^n=f_a^n(x_q^{n-1};W^n,b^n)$;
\STATE \hspace{2.0cm} Calculate the ReSU/STSU activation $x_q^n$ based on Eq.~\eqref{eq:resu}/\eqref{eq:masu};
\STATE \hspace{2.0cm} Adjust the firing threshold $V_{th}^n$ in $n$ th layer based on Eq.~\eqref{eq:ata};
\STATE \hspace{1.0cm} \textbf{For} $n=L$ to $1$
\STATE \hspace{1.5cm} Calculate $\frac{\partial \mathcal{L}}{\partial W^n}$ and $\frac{\partial \mathcal{L}}{\partial b^n}$ based on Eq.~\eqref{eq:ste1},~\eqref{eq:ste2} or Eq.~\eqref{eq:stsu1},~\eqref{eq:stsu2};
\STATE \hspace{1.5cm} Update: $W^n \leftarrow W^n -\beta\frac{\partial \mathcal{L}}{\partial W^n},b^n \leftarrow b^n -\beta\frac{\partial \mathcal{L}}{\partial b^n}$;
\STATE \textbf{Return} $\{W^n, b^n, V_{th}^n\}_{n=1}^L$.
\end{algorithmic}
\label{algo:1}
\end{algorithm}
\section{Experiments}
\subsection{Experiment Implementation}
We validate the SNN2ANN with VGG-13 and ResNet-17 on the CIFAR10 \cite{krizhevsky2009learning}, CIFAR100 \cite{krizhevsky2009learning}, and Tiny-ImageNet \cite{le2015tiny} datasets in two aspects: \textbf{a). Classification performance}; \textbf{b). The efficiencies of training and testing.}  Details of the experiment settings are given in the following subsections.

\subsection{Experiment Settings}
\subsubsection{Datasets and Network Architectures}
Details of the datasets are presented in Table~\ref{tab:dataset}, and the network architectures for experiments are given in Table~\ref{tab:arch}. In Table~\ref{tab:arch}, $k$ is the kernel size, $c$ is the dimensions of the channel, and $s$ denotes the stride. $(M)$ denotes the max-pooling for Tiny-ImageNet. The stride of the $10$-th and $14$-th convolution layers of ResNet-17 is set to $2$. For Tiny-ImageNet, we additionally set $\text{stride}=2$ to the $6$-th layer of ResNet-17. For the STBP-based VGG-13, we apply the spiking max-pooling layer \cite{Fang_2021_ICCV} to enlarge the receptive field and utilize the BN layer to remove the deviation.
\begin{table}[h]
  \centering
  \renewcommand{\arraystretch}{1.1}
  \caption{Benchmark datasets.}
  \scalebox{1.0}{
    \begin{tabular}{|c|c|c|c|}
    \hline
    \textbf{Dataset} & \textbf{CIFAR10} & \textbf{CIFAR100} & \textbf{Tiny-ImageNet} \\
    \hline
    \textbf{Size} & $32\times32\times3$ & $32\times32\times3$ & $64\times64\times3$ \\
    \hline
    \textbf{Training Samples} & 50,000 & 50,000 & 100,000 \\
    \hline
    \textbf{Testing Samples} & 10,000 & 10,000 & 10,000 \\
    \hline
    \textbf{Category} & 10    & 100   & 200 \\
    \hline
    \end{tabular}%
    }
  \label{tab:dataset}%
\end{table}%
\begin{table}[h]
  \centering
  \renewcommand{\arraystretch}{1.0}
  \caption{Network architectures for experiments. }
  \scalebox{1.0}{
    \begin{tabular}{|c|c|}
    \hline
    \textbf{Network} & \textbf{Architecture} \\
    \hline
    {\textbf{VGG-13}} & $[k3c64s1]\times 2$ -$[k3c128s1]\times 2$-$(M)$-$[k3c256s1]\times 3$-$M$  \\
          & -$[k3c512s1]\times 3$-$M$-$[k3c512s1]\times 3$-$M$-$FC$ \\
    \hline
    {\textbf{ResNet-17}} & $k3c64s1-[k3c64$-$k3c64]\times 2$-$[k3c128$-$k3c128]\times 2$ \\
          & -$[k3c256$-$k3c256]\times 2$-$[k3c512$-$k3c512]\times 2$-$FC$ \\
    \hline
    \end{tabular}%
    }
  \label{tab:arch}%
\end{table}%

\subsubsection{Pooling Layer in SNN2ANN-based VGG}
Max-pooling \cite{DBLP:journals/corr/SimonyanZ14a} is widely used in CNN as it naturally picks the prominent features among the inputs. Since the spiking trains determine the max spike rate, most current SNNs apply the average pool on the spiking outputs to expand the receptive field \cite{wu2019direct,10.3389/fnins.2019.00095,DBLP:conf/ijcai/DingY0H21}. Unlike the existing spiking pooling layer, the SNN branch of SNN2ANN operates the max-pooling on the convolution output at each time step, and the pooling feature maps are fed to the IF neuron to generate the spikes. The max-pooling selects the max convolution results as the MP without introducing any parameters and MAC operations. It is reasonable to expand the receptive field of SNN.
\subsubsection{Training Setting}
We apply the Cross-Entropy Loss to guide the learning of networks. Adam \cite{DBLP:journals/corr/KingmaB14} optimizer with an initial learning rate of $0.001$ is used to minimize the classification errors. We train the ANN models for 200 epochs on all datasets.  Details of the SNN training setting for each dataset are presented in Table~\ref{tab:setting}. We decrease the learning rate by $90\%$ for all models at the $100$-th, $150$-th, $175$-th, $300$-th, $350$-th,  and $375$-th epochs. For VGG-13 on CIFAR10/Tiny-ImageNet, we reset the learning rate to 0.001 at the $200$-th epoch. Referring to \cite{wu2019direct,DBLP:conf/aaai/YanZW21,xiao2021training}, we directly feed the original RGB image into the network, and the first spiking layer is regarded as the spiking encoder. For ATA, we set $\tau=\alpha=0.1$. For BN, we set $\alpha=0.1$. The experiments are conducted on the PyTorch platform. The GPUs used in training are 8 GeForce RTX 8000.
\begin{table}[htbp]
  \centering
  \caption{Training setting.}
  \scalebox{1.0}{
    \begin{tabular}{|c|c|c|c|c|}
    \hline
    \textbf{Dataset} & \textbf{Model} & \textbf{Time Steps} & \textbf{Epochs} & \textbf{Batch Size} \\
    \hline
    \multirow{2}{*}{CIFAR10} & VGG-13 & 5     & 400   & 512 \\
\cline{2-5}          & ResNet-17 & 5     & 200   & 512 \\
    \hline
    \multirow{2}{*}{CIFAR100} & VGG-13 & 4     & 200   & 512 \\
\cline{2-5}          & ResNet-17 & 5     & 200   & 512 \\
    \hline
    \multirow{2}{*}{Tiny-ImageNet} & VGG-13 & 3     & 400   & 1024 \\
\cline{2-5}          & ResNet-17 & 5     & 200   & 1024 \\
    \hline
    \end{tabular}%
    }
  \label{tab:setting}%
\end{table}%
\subsection{Classification Performance}
\begin{table}[t]
  \centering
  \renewcommand{\arraystretch}{1.2}
  \caption{Comparison between SNN2ANN and SOTA's models. \textbf{``A2S''} and \textbf{``S2A-ReSU/STSU''} denote ANN2SNN and SNN2ANN, respectively. }
  \scalebox{0.8}{
    \begin{tabular}{|c|c|c|c|c|}
    \hline
    \multicolumn{5}{|c|}{\textbf{CIFAR10}} \\
    \hline
    \textbf{Reference} & \textbf{Method} & \textbf{Arch} & \textbf{Acc (\%)} & \textbf{Steps} \\
    \hline
    CVPR2020\cite{DBLP:conf/cvpr/0006S020} & A2S & VGG-16 & 93.63 & 2048 \\
    \hline
    ICLR2021\cite{DBLP:journals/corr/abs-2103-00476} & A2S & ResNet-20 & 93.58 & 400-600 \\
    \hline
    AAAI2021\cite{DBLP:conf/aaai/YanZW21} & A2S & VGG-$\ast$ & 94.16 & 600 \\
    \hline
    AAAI2022\cite{DBLP:journals/corr/abs-2202-01440} & A2S & ResNet-20 & 92.75 & $\geq$512 \\
    \hline
    NMI2021\cite{DBLP:journals/natmi/StocklM21} & A2S & ResNet-20 & 91.45 & 200 \\
    \hline
    TNNLS2021\cite{9492305} & Tandem & CIFARNET & 90.98 & 8 \\
    \hline
    TNNLS2021\cite{DBLP:journals/corr/abs-2008-03658} & A2S\&BP & VGG-16 & 92.7 & 5 \\
    \hline
    TNNLS2022\cite{DBLP:journals/tnn/YuMSZDT22} & A2S & VGG-16 & 93.29 & 1100 \\
    \hline
    IJCAI2021\cite{DBLP:conf/ijcai/DingY0H21} & A2S & PreActResNet-18 & 93.41 & 256 \\
    \hline
    ICCV2021\cite{garg2021dct} & BP    & DCT+VGG9 & 89.94 & 48 \\
    \hline
    ICCV2021\cite{Fang_2021_ICCV} & BP    & 8-Layer CNN & 93.5  & 8 \\
    \hline
    AAAI2021\cite{DBLP:conf/aaai/Zheng00HL21} & BP    & ResNet-19 & 93.16 & 6 \\
    \hline
    ICLR2022\cite{DBLP:journals/corr/abs-2202-11946} & BP    & ResNet-19 & 94.50 & 6 \\
    \hline
    NeurIPS2021\cite{li2021differentiable} & BP    & ResNet-18 & 94.25 & 6 \\
    \hline
    NeurIPS2021\cite{xiao2021training} & IDE-BP & CIFARNET-F & 92.52 & 100 \\
    \hline
    \multirow{4}{*}{\textbf{This Work}} & S2A & VGG-13 & 92.62$\pm$0.03 & \textbf{5} \\
\cline{3-5}          & -ReSU  & ResNet-17 & 92.84$\pm$0.03& \textbf{5} \\
\cline{2-5}          & S2A & VGG-13 & 92.18$\pm$0.07 & \textbf{5} \\
\cline{3-5}          & -STSU  & ResNet-17 & 92.75$\pm$0.03 & \textbf{5} \\
    \hline
    \multicolumn{5}{|c|}{\textbf{CIFAR100}} \\
    \hline
    \textbf{Reference} & \textbf{Method} & \textbf{Arch} & \textbf{Acc (\%)} & \textbf{Steps} \\
    \hline
    CVPR2020\cite{DBLP:conf/cvpr/0006S020} & A2S & VGG-16 & 70.93 & 2048 \\
    \hline
    ICLR2021\cite{DBLP:journals/corr/abs-2103-00476} & A2S & VGG-16 & 70.55 & 400-600 \\
    \hline
    AAAI2021\cite{DBLP:conf/aaai/YanZW21} & A2S & VGG-$\ast$ & 71.52 & 600 \\
    \hline
    AAAI2022\cite{DBLP:journals/corr/abs-2202-01440} & A2S & ResNet-20 & 70.53 & $\geq$ 512 \\
    \hline
    TNNLS2021\cite{DBLP:journals/corr/abs-2008-03658} & A2S\&BP & VGG-16 & 69.67 & 5 \\
    \hline
    IJCAI2021\cite{DBLP:conf/ijcai/DingY0H21} & A2S & PreActResNet-18 & 75.1  & 256 \\
    \hline
    ICCV2021\cite{garg2021dct} & BP    & DCT+VGG-11 & 68.83 & 48 \\
    \hline
    ICLR2022\cite{DBLP:journals/corr/abs-2202-11946} & BP    & ResNet-19 & 74.72 & 6 \\
    \hline
    NeurIPS2021\cite{xiao2021training} & IDE-BP & CIFARNET-F & 73.07 & 100 \\
    \hline
    NeurIPS2021\cite{li2021differentiable} & BP    & ResNet-18 & 74.24 & 6 \\
    \hline
    \multirow{4}{*}{\textbf{This Work}} & S2A & VGG13 & 71.10$\pm$0.06 & \textbf{4} \\
\cline{3-5}          & -ReSU  & ResNet17 & 72.92$\pm$0.14 & \textbf{5} \\
\cline{2-5}          & S2A & VGG-13 & 68.96$\pm$0.24 & \textbf{4} \\
\cline{3-5}          & -STSU  & ResNet-17 & 73.26$\pm$0.20 & \textbf{5} \\
    \hline
    \multicolumn{5}{|c|}{\textbf{Tiny-ImageNet}} \\
    \hline
    \textbf{Reference} & \textbf{Training} & \textbf{Arch} & \textbf{Acc (\%)} & \textbf{Steps} \\
    \hline
    FRONT NEUSCI2019\cite{10.3389/fnins.2019.00095}  & A2S & VGG-16 & 48.6  & 2500 \\
    \hline
    ICCV2021\cite{kundu2021spike} & A2S\&BP & VGG-16 & 51.92 & 150 \\
    \hline
    ICCV2021\cite{garg2021dct} & DCT-SNN & VGG-16 & 52.43 & 125 \\
    \hline
    \multirow{4}{*}{\textbf{This Work}} & S2A & VGG-13 & \textbf{54.91$\pm$0.20} & \textbf{3} \\
\cline{3-5}          & -ReSU  & ResNet-17 & \textbf{56.25$\pm$0.04} & \textbf{5} \\
\cline{2-5}          & S2A & VGG-13 & \textbf{54.33$\pm$0.05} & \textbf{3} \\
\cline{3-5}          & -STSU  & ResNet-17 & \textbf{56.91$\pm$0.07} & \textbf{5} \\
    \hline
    \end{tabular}%
    }
  \label{tab:sota}%
\end{table}%
We first compare the performance of our models with the state-of-the-art (SOTA) SNNs, and the results are given in Table~\ref{tab:sota}. Each accuracy result in Table~\ref{tab:sota} is the average of the $5$ best running performance. Both SNN2ANN-based VGG-13 and ResNet-17 achieve complete performance with the SOTA models. For CIFAR10 and CIFAR100, the SNN2ANN-based ResNet-17 achieves accuracy with $92.84\%$ and $73.26\%$, respectively. \cite{DBLP:journals/corr/abs-2202-11946} achieves $94.5\%$ accuracy on CIFAR10, which outperforms other BP-based models. However, such a model is trained by BPTT, requiring many training resources. The ANN2SNN \cite{DBLP:conf/ijcai/DingY0H21} achieves the best accuracy with $75.1\%$ on CIFAR100, but the inference time steps are $51\times$ larger than our models. For Tiny-ImageNet, our VGG-13 and ResNet-17 achieve $54.81\%$ and $56.25\%$ accuracies, outperforming the comparison methods. The SNN2ANN models are inference-efficient and only require $3\sim 5$ inference steps.

Using the VGG-13 and ResNet-17, we compare our SNN2ANN with the STBP-based \cite{wu2019direct} models with IF and Parametric LIF (PLIF) \cite{Fang_2021_ICCV} neurons and an ANN2SNN model: RNL-RIL \cite{DBLP:conf/ijcai/DingY0H21}. The comparison results are shown in Table~\ref{tab:same_cmp}. Both SNN2ANN and STBP-PLIF models achieve complete performance, and the SNN2ANN models have the least spike activities and time steps. The RNL-RIL outperforms other models. However, they require hundred-time steps and many spike numbers for high accuracy. Benefiting from the PLIF neuron introducing the trainable decay factors, the STBP-PLIF outperforms better than the STBP-IF models. The STBP-IF-Based models suffer the over-fitting problem during training, resulting in unsatisfactory performance in many cases. Compared with the STBP-based training, the SNN2ANN models outperform the STBP-based models for all datasets. In addition, the STBP-PLIF models requires more inference steps than SNN2ANN models to achieve comparable performance.
\begin{table*}[!htbp]
  \centering
  \renewcommand{\arraystretch}{1.2}
  \caption{Classification Performance of RNL-RIL, STBP, and SNN2ANN (\textbf{S2A-\#}) with the same architectures.}

  \scalebox{1.0}{
    \begin{tabular}{|c|c|c|c|c|c|c|c|c|c|}
    \hline
    \multirow{2}[4]{*}{\textbf{Method}} & \multicolumn{3}{c|}{\textbf{CIFAR10}} & \multicolumn{3}{|c|}{\textbf{CIFAR100}} & \multicolumn{3}{c|}{\textbf{Tiny-ImageNet}} \\
\cline{2-10}          & \textbf{Acc (\%)} & \textbf{Steps} & \textbf{Spikes/Image} & \textbf{Acc (\%)} & \textbf{Steps} & \textbf{Spikes/Image} & \textbf{Acc (\%)} & \textbf{Steps} & \textbf{Spikes/Image} \\
    \hline
    \multicolumn{10}{|c|}{\textbf{VGG-13}} \\
    \hline
    \textbf{ANN} & 94.05  & -     & -     & 73.01  & -     & -     & 56.19  & -     & - \\
    \hline
    \textbf{RNL-RIL} & 92.50  & 250   & $4.24\times 10^6$ & \textbf{72.90} & 250   & $5.94\times 10^6$ & \textbf{56.10} & 250   & $2.03\times 10^7$ \\
    \hline
    \multirow{2}{*}{\textbf{STBP-IF}} & 84.20  & 5     & $1.11\times 10^6$ & 57.77  & 4     & $1.35\times 10^6$ & \multirow{2}{*}{54.53} & \multirow{2}{*}{3} & \multirow{2}{*}{$2.17\times 10^6$} \\
\cline{2-7}          & 71.54  & 8     & $2.01\times 10^6$ & 39.86  & 8     & $2.69\times 10^6$ &       &       &  \\
    \hline
    \textbf{STBP-PLIF} & 91.63  & 5     & $9.67\times 10^5$ & 70.94  & 4     & $1.05\times 10^6$ & 53.08  & 3     & $1.74\times 10^6$ \\
    \hline
    \textbf{S2A-ReSU} & \textbf{92.62} & \textbf{5}     & $1.68\times 10^6$ & \textbf{71.10} & \textbf{4}     & \textbf{$\mathbf{6.69\times 10^5}$} & \textbf{54.91} & \textbf{3}     & \textbf{$\mathbf{1.02\times 10^6}$} \\
    \hline
    \textbf{S2A-STSU} & \textbf{92.18} & \textbf{5}     & \textbf{$\mathbf{4.52\times 10^5}$} & 68.96  & \textbf{4}     & \textbf{$\mathbf{6.18\times 10^5}$} & 54.33  & \textbf{3}     & \textbf{$\mathbf{1.11\times 10^6}$} \\
    \hline
    \multicolumn{10}{|c|}{\textbf{ResNet-17}} \\
    \hline
    \textbf{ANN} & 93.87  & -     & -     & 73.64  & -     & -     & 56.04  & -     & - \\
    \hline
    \multirow{2}{*}{\textbf{STBP-IF}} & 83.01  & 5     & $1.42\times 10^6$ & 55.50  & 5     & $1.79\times 10^6$ & \multirow{2}{*}{39.22 } & \multirow{2}{*}{5} & \multirow{2}{*}{$3.59\times 10^6$} \\
\cline{2-7}          & 76.23  & 8     & $2.34\times 10^6$ & 42.90  & 8     & $2.90\times 10^6$ &       &       &  \\
    \hline
    \multirow{2}{*}{\textbf{STBP-PLIF}} & 91.33  & 5     & $1.43\times 10^6$ & 70.60  & 5     & $1.50\times 10^6$ & \multirow{2}{*}{56.84} & \multirow{2}{*}{5} & \multirow{2}{*}{$3.13\times 10^6$} \\
\cline{2-7}          & 92.32  & 8     & $2.34\times 10^6$ & 71.58 & 8     & $2.43\times 10^6$ &       &       &  \\
    \hline
    \textbf{S2A-ReSU} & \textbf{92.84} & \textbf{5}     & \textbf{$\mathbf{7.42\times 10^5}$} & 72.92  & \textbf{5}     & \textbf{$\mathbf{9.88\times 10^5}$} & 56.25  & \textbf{5}     & \textbf{$\mathbf{1.49\times 10^6}$} \\
    \hline
    \textbf{S2A-STSU} & \textbf{92.75} & \textbf{5}     & \textbf{$\mathbf{6.58\times 10^5}$} & \textbf{73.26}  & \textbf{5}     & \textbf{$\mathbf{8.10\times 10^5}$} & \textbf{56.91} & \textbf{5}     & $\mathbf{1.71\times 10^6}$ \\
    \hline
    \end{tabular}%
    }
  \label{tab:same_cmp}%
\end{table*}%
\subsection{Efficiencies}
In this part, we analyze the training efficiency of the SNN2ANN framework. There are two aspects to discuss: \textbf{1) Training efficiency} and \textbf{2) Inference efficiency.} We compare our SNN2ANN models with the RNL-RIL, the STBP-IF, and STBP-PLIF-based models. For the STBP-based models, we apply the Spikingjelly \cite{SpikingJelly} for model implementation.
\subsubsection{Training Efficiency}
\begin{figure*}[t]
\centering
\renewcommand{\arraystretch}{1.3}
\subfigure[]{
\includegraphics[width=0.3\textwidth]{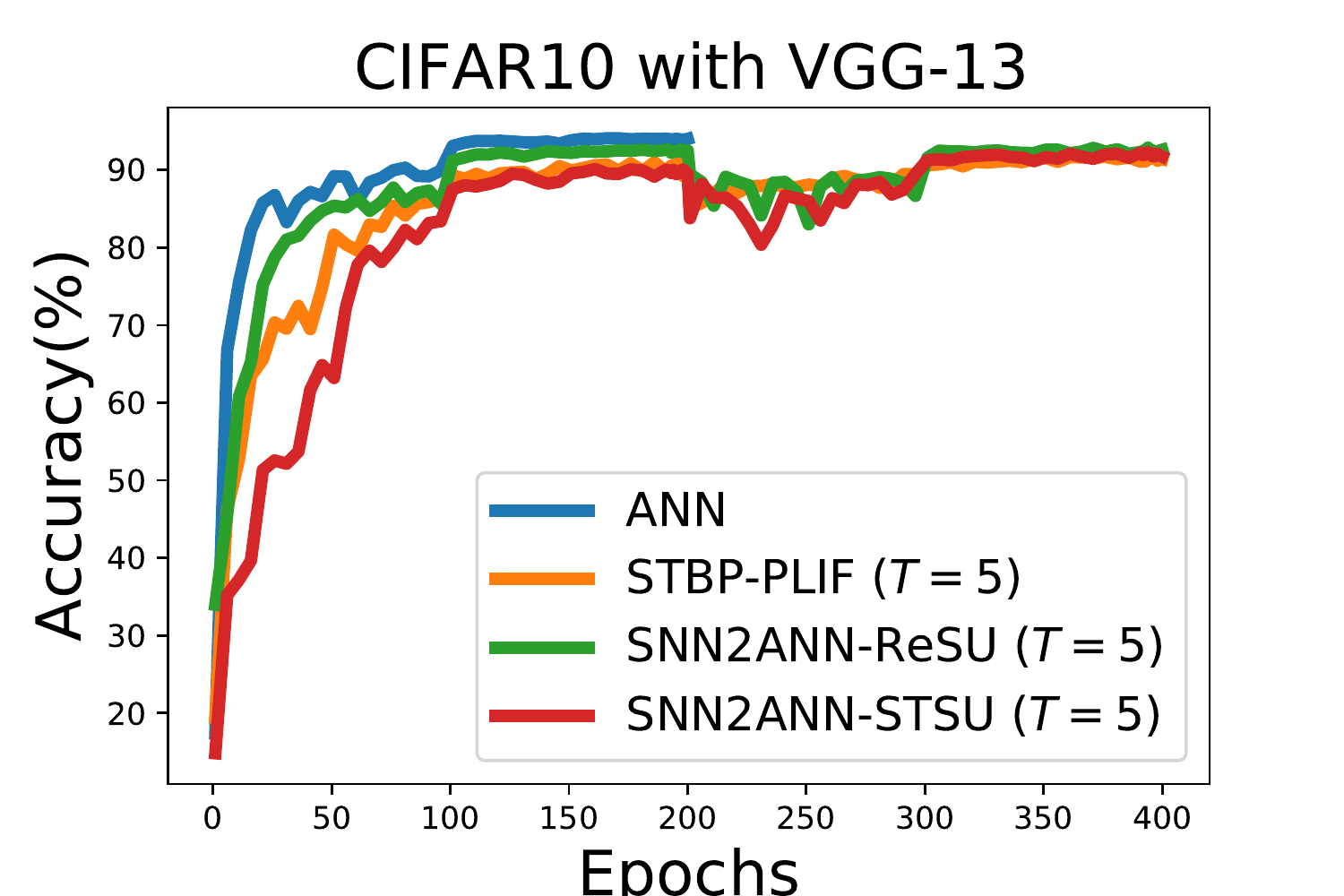}}
\subfigure{
\includegraphics[width=0.3\textwidth]{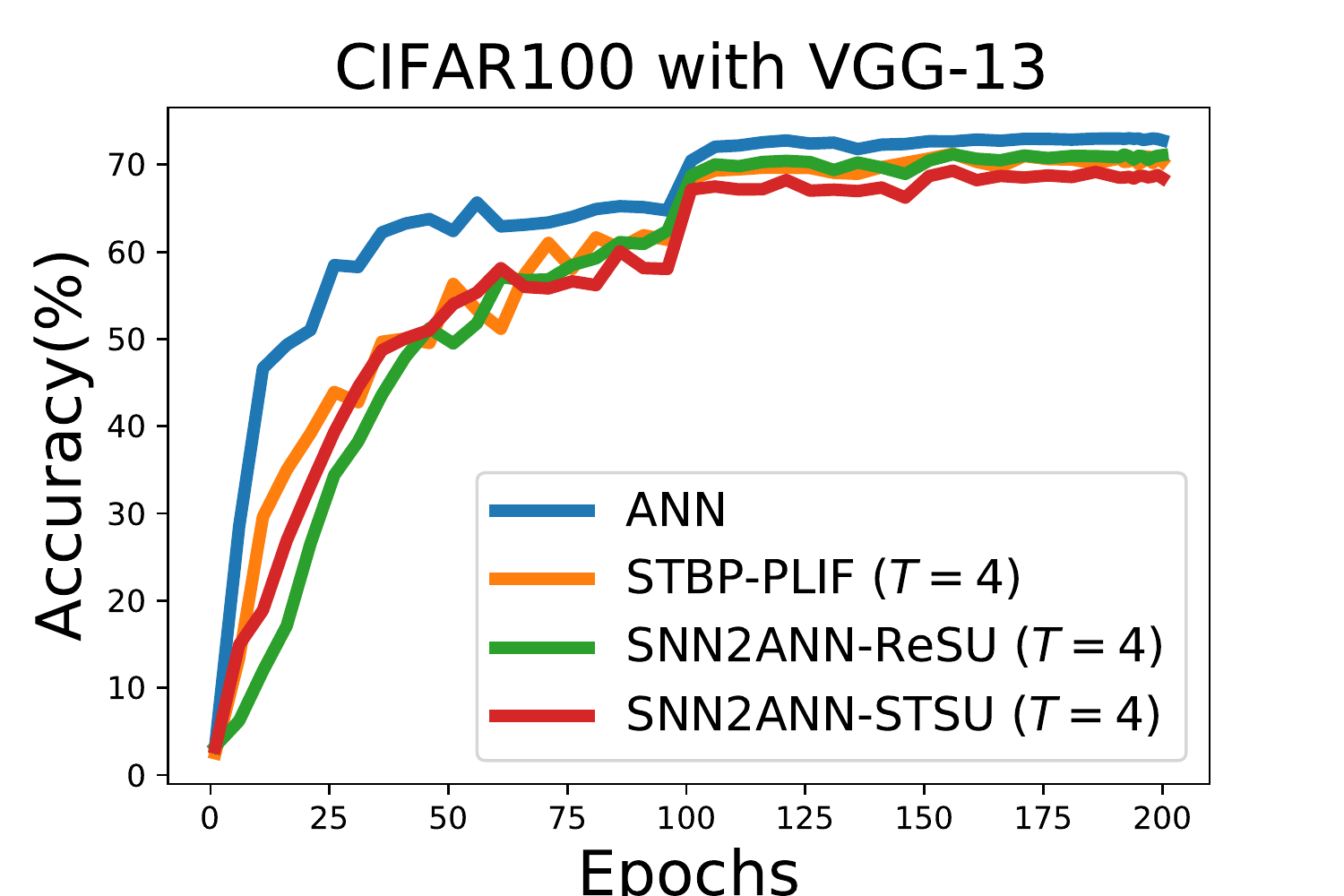}}
\subfigure{
\includegraphics[width=0.3\textwidth]{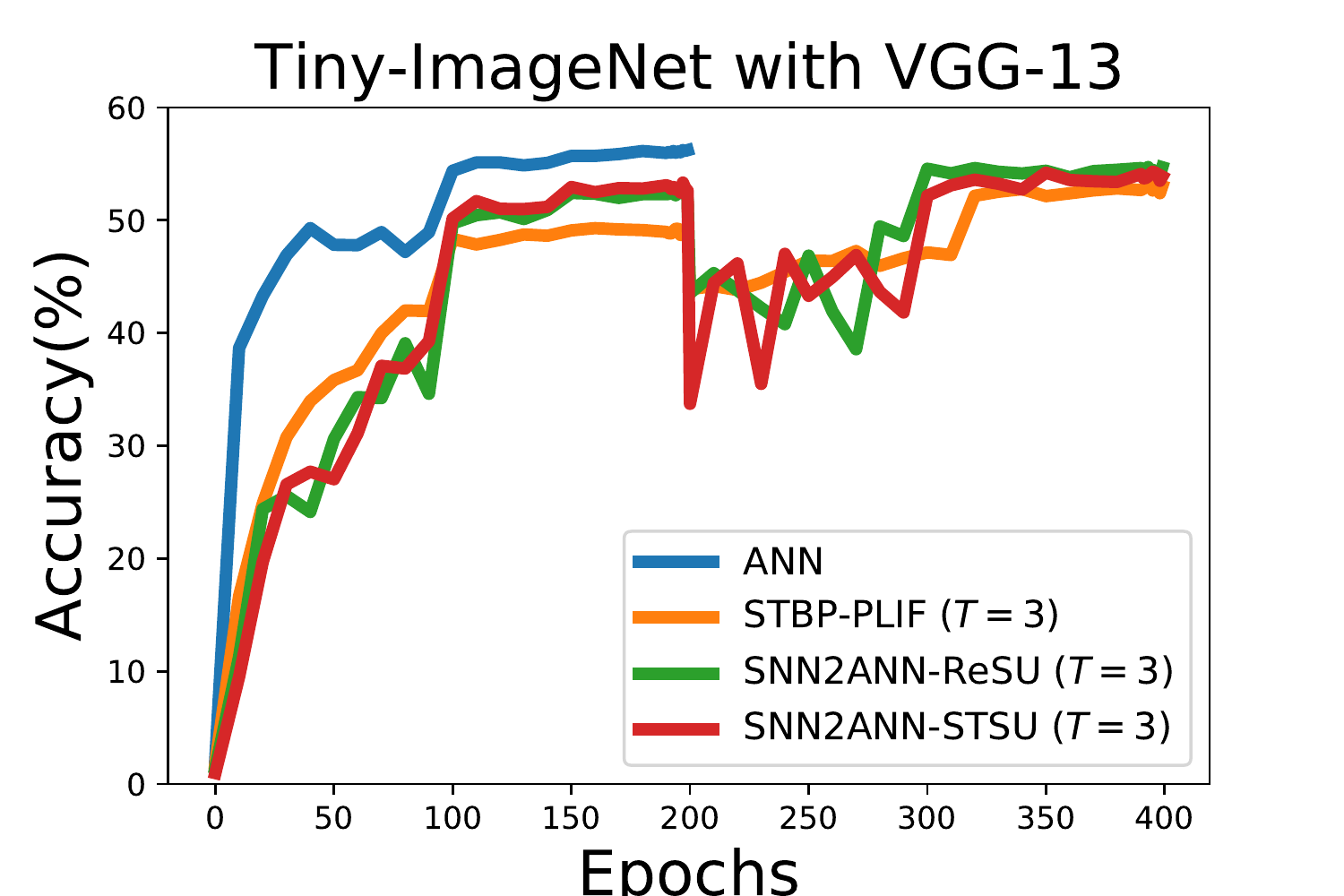}}
\subfigure{
\includegraphics[width=0.3\textwidth]{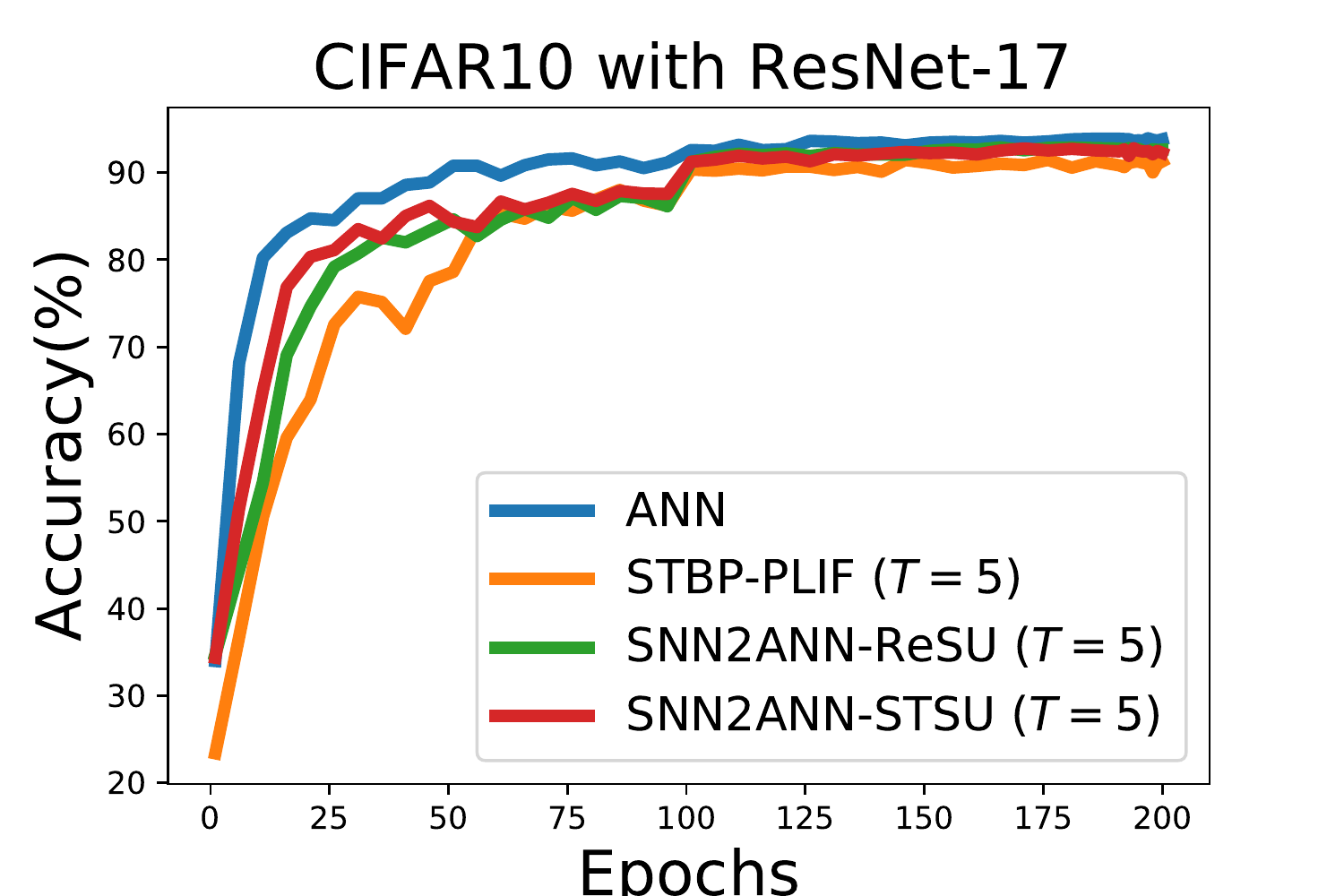}}
\subfigure{
\includegraphics[width=0.3\textwidth]{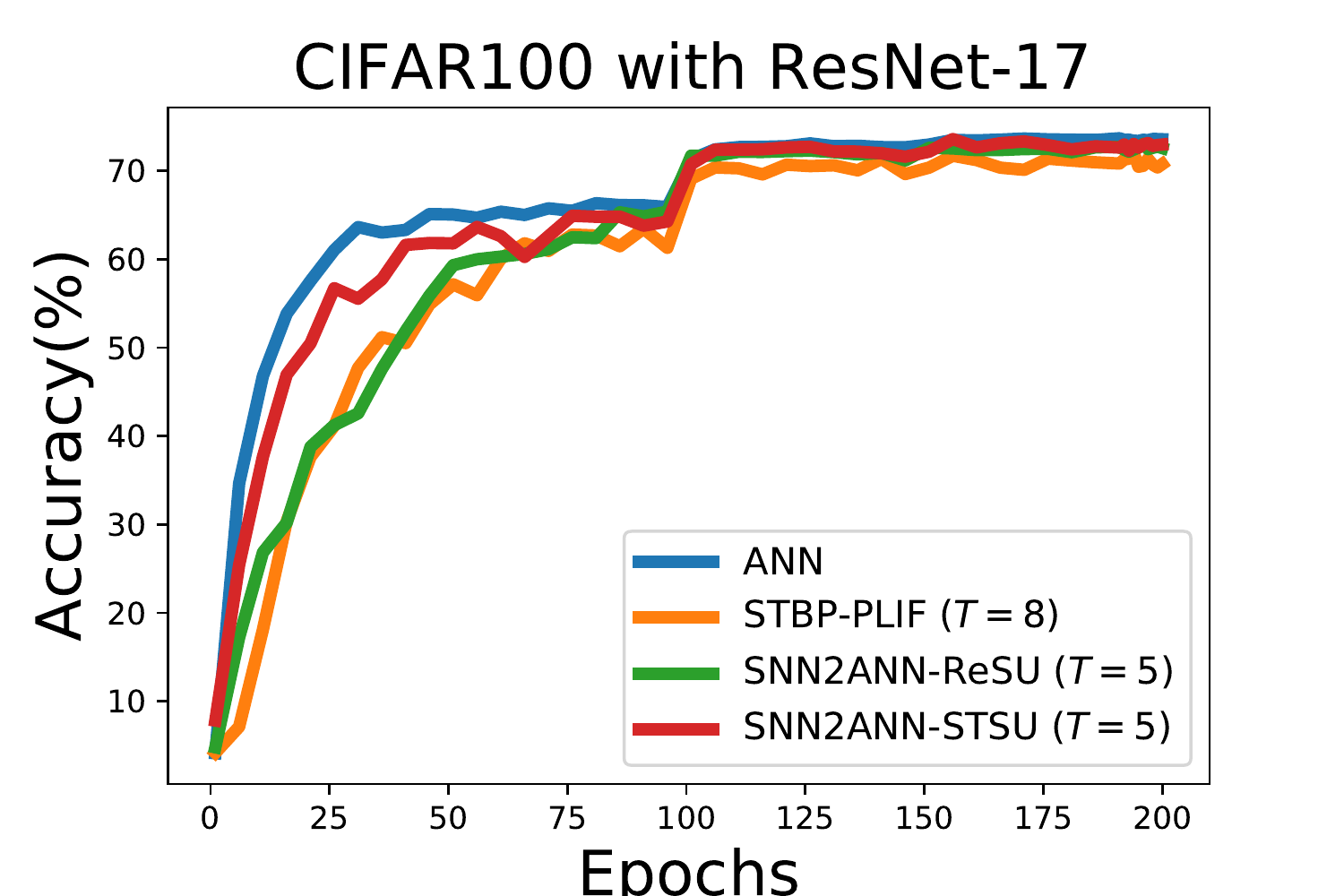}}
\subfigure{
\includegraphics[width=0.3\textwidth]{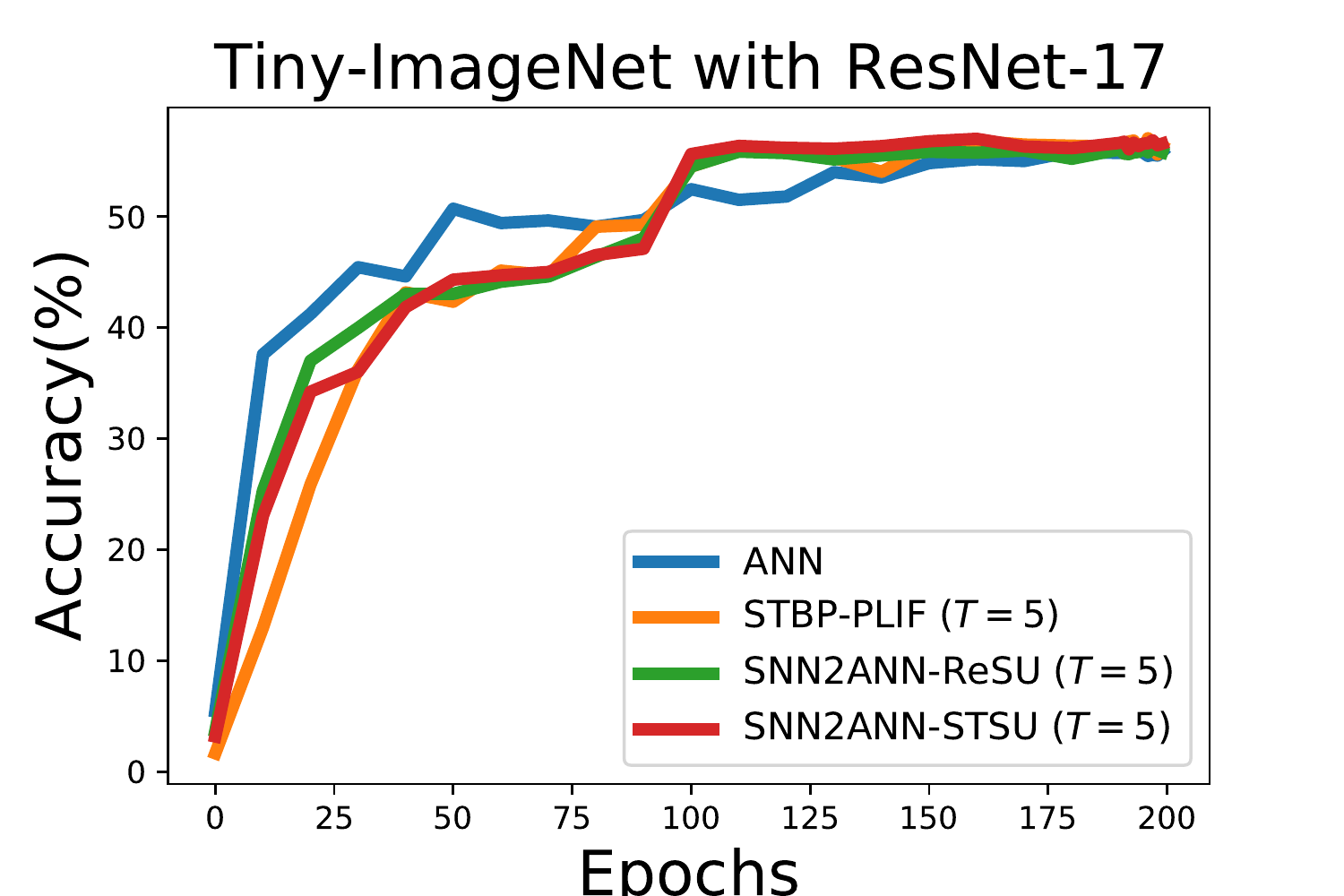}}
\caption{The changing accuracy of SNN2ANN and STBP-PLIF on the benchmark datasets.}
\label{fig:curve_cmp}
\end{figure*}

The training efficiencies are evaluated based on the convergence, training time, and GPU memory costs. Fig.~\ref{fig:curve_cmp} displays the changing accuracies of ANN, SNN2ANN, and STBP-PLIF-based models on the benchmark datasets. The blue curves denote the ANN, the orange curves are the STBP-PLIF models, and the green and red curves are the SNN2ANN models.

From the view of training epochs, the ANN models converge faster than SNN2ANN and the STBP-based models. Both SNN2ANN and STBP-based models show a similar convergence tendency after $100$ epoch training. Therefore, the convergence of iterations for SNN2ANN is similar to the STBP.

\begin{table}[htbp]
  \centering
  \renewcommand{\arraystretch}{1.0}
  \caption{Training costs comparisons between STBP and SNN2ANN (\textbf{S2A-ReSU/STSU}).}
  \scalebox{0.9}{
    \begin{tabular}{|c|c|c|}
    \hline
    \textbf{Method} & \textbf{Train Time (sec/epoch) $\downarrow$} & \textbf{GPU Mem (MiB)} $\downarrow$ \\
    \hline
    \multicolumn{3}{|c|}{\textbf{VGG-13 }} \\
    \hline
    \multicolumn{3}{|c|}{\textbf{CIFAR10}} \\
    \hline
    STBP-IF (T=5) & 55.23 & $7.48\times 10^4$ \\
    \hline
    STBP-PLIF (T=5) & 57.88 & $9.31\times 10^4$ \\
    \hline
    \textbf{S2A-ReSU (T=5)} & \textbf{24.96} & \textbf{$\mathbf{3.52\times 10^4}$} \\
    \hline
    \textbf{S2A-STSU (T=5)} & \textbf{24.73} & \textbf{$\mathbf{3.04\times 10^4}$} \\
    \hline
    \multicolumn{3}{|c|}{\textbf{CIFAR100}} \\
    \hline
    STBP-IF (T=4) & 44.31 & $6.37\times 10^4$ \\
    \hline
    STBP-PLIF (T=4) & 47.64 & $7.88\times 10^4$ \\
    \hline
    \textbf{S2A-ReSU} (T=4) & \textbf{25.12} & \textbf{$\mathbf{3.48\times 10^4}$} \\
    \hline
    \textbf{S2A-STSU (T=4)} & \textbf{21.67} & \textbf{$\mathbf{3.18\times 10^4}$} \\
    \hline
    \multicolumn{3}{|c|}{\textbf{Tiny-ImageNet}} \\
    \hline
    STBP-IF (T=3) & 89.80  & $1.89\times 10^5$ \\
    \hline
    STBP-PLIF (T=3) & 90.53 & $1.91\times 10^5$ \\
    \hline
    \textbf{S2A-ReSU (T=3)} & \textbf{53.55} & \textbf{$\mathbf{7.94\times 10^4}$} \\
    \hline
    \textbf{S2A-STSU (T=3)} & \textbf{52.98} & \textbf{$\mathbf{6.95\times 10^4}$} \\
    \hline
    \multicolumn{3}{|c|}{\textbf{ResNet-17 }} \\
    \hline
    \multicolumn{3}{|c|}{\textbf{CIFAR10}} \\
    \hline
    STBP-IF (T=5) & 47.82 & $5.85\times 10^4$ \\
    \hline
    STBP-PLIF (T=5) & 50.90  & $7.13\times 10^4$ \\
    \hline
    STBP-PLIF (T=8) & 79.29 & $1.04\times 10^5$ \\
    \hline
    \textbf{S2A-ReSU} (T=5) & \textbf{31.86} & \textbf{$\mathbf{3.25\times 10^4}$} \\
    \hline
    \textbf{S2A-STSU (T=5)} & \textbf{31.74} & \textbf{$\mathbf{2.86\times 10^4}$} \\
    \hline
    \multicolumn{3}{|c|}{\textbf{CIFAR100}} \\
    \hline
    STBP-IF (T=5) & 48.71 & $5.93\times 10^4$ \\
    \hline
    STBP-PLIF (T=5) & 52.23 & $7.23\times 10^4$ \\
    \hline
    STBP-PLIF (T=8) & 84.08 & $1.05\times 10^5$ \\
    \hline
    \textbf{S2A-ReSU} (T=5) & \textbf{32.17} & \textbf{$\mathbf{3.26\times 10^4}$} \\
    \hline
    \textbf{S2A-STSU} (T=5) & \textbf{31.73} & \textbf{$\mathbf{2.87\times 10^4}$} \\
    \hline
    \multicolumn{3}{|c|}{\textbf{Tiny-ImageNet}} \\
    \hline
    STBP-IF (T=5) & 114.54 & $1.76\times 10^5$ \\
    \hline
    STBP-PLIF (T=5) & 120.89 & $2.20\times 10^5$ \\
    \hline
    \textbf{S2A-ReSU (T=5)} & \textbf{71.58} & \textbf{$\mathbf{8.51\times 10^4}$} \\
    \hline
    \textbf{S2A-STSU (T=5)} & \textbf{70.01} & \textbf{$\mathbf{7.43\times 10^4}$} \\
    \hline
    \end{tabular}%
    }
  \label{tab:train_eff}%
\end{table}%

To further analyze the training efficiency of the SNN2ANN, Table~\ref{tab:train_eff} presents the GPU memory costs and average training time in an epoch of SNN2ANN and STBP. Compared with the STBP, our SNN2ANN trains the SNNs in less training time and GPU memory costs. We first analyze the training cost under the same time steps setting. For the VGG-13, our SNN2ANN trains the networks with $0.448\times$ $\sim$ $0.596\times$ training time and $0.369\times$ $\sim$ $0.546\times$ memory cost of the STBP-IF models, $0.427\times$ $\sim$ $0.592\times$ training time and $0.327\times$ $\sim$ $0.441\times$ memory cost of the STBP-PLIF models. For the ResNet-17, the training time and memory cost of our SNN2ANN are $0.592\times$ $\sim$ $0.666\times$ and $0.423\times$ $\sim$ $0.556\times$ of the STBP-IF models, $0.579\times$ $\sim$ $0.626\times$ and $0.337\times$ $\sim$ $0.455\times$ of the STBP-PLIF models. In addition, based on Table~\ref{tab:same_cmp}, we can find that our SNN2ANN models outperform the STBP-based models under the same time setting in most cases and always achieve the least spike activities. It demonstrates that our SNN2ANN can train the SNNs in a fast and low memory cost way.

Based on Tables ~\ref{tab:same_cmp} and~\ref{tab:train_eff}, we analyze the training cost when the performance of SNN2ANN and STBP-based models are similar. For CIFAR10, our SNN2ANN-ReSU ResNet-17 achieves $92.84\%$ with $5$ steps, while the STBP-PLIF ResNet-17 requires $8$ steps to achieve $92.32\%$ accuracy. The large time steps of the STBP-PLIF ResNet-17 require an extensive memory cost and slow down the training speed. Our SNN2ANN-ReSU trains the ResNet-17 with $0.402\times$ training time and $0.313\times$ memory cost of the STBP-PLIF. For CIFAR100, our SNN2ANN-STSU ResNet-17 achieve $73.26\%$ accuracy under $0.377\times$ training time, and $0.27\times$ GPU memory cost of the STBP-PLIF ResNet-17 ($T=8$). Both SNN2ANN-STSU and STBP-PLIF ResNet-17 achieve similar performance under the same time steps on Tiny-ImageNet. However, the training time and memory cost of SNN2ANN-STSU are $0.579\times$ and $0.386\times$ of the STBP-PLIF, respectively.
\subsubsection{Inference Efficiency}
We analyze the inference efficiencies in 3 aspects: \textbf{i) Inference time, ii) Spike activities,} and \textbf{iii) Energy efficiency}.\\
\noindent\textbf{i) Inference time:} Table~\ref{tab:same_cmp} gives the inference time of the SNN2ANN, STBP, and RNL-RIL models. Both SNN2ANN and STBP achieve a shorter inference time than the ANN2SNN models. The time steps of RNL-RIL models are $245$, $246$, and $247$ steps more than our SNN2ANN VGG-13 on CIFAR10, CIFAR100, and Tiny-ImageNet, respectively. The RNL-RIL-based VGG-13 performs the best accuracy among other VGG-13-based models on CIFAR100 and Tiny-ImageNet. However, the spike numbers of the RNL-RIL-based VGG-13 are $2.51\times$ $\sim$ $19.8\times$ and $9.36\times$ $\sim$ $18.25\times$ of our SNN2ANN-ReSU and STSU VGG-13, respectively. Then, we analyze the efficiency of the inference time based on the SNN2ANN and STBP-based models that have similar accuracy performance. The inference steps of SNN2ANN models for comparable accuracy are less or equal to the STBP-based models for all datasets. For CIFAR10, the SNN2ANN-ReSU ResNet-17 uses $5$ steps for inference and outperforms a $0.52\%$ accuracy value than the STBP-PLIF ResNet-17, which time steps are $8$. For CIFAR100, our SNN2ANN-STSU ResNet-17 achieves $73.26\%$ under $0.625\times$ time steps and $0.33\times$ spike activities of the STBP-PLIF ResNet-17.

\noindent\textbf{ii) Spike activities:} Combining the accuracy and spikes number in Table~\ref{tab:same_cmp}, we can find that our SNN2ANN models achieve complete accuracy with fewer spikes activities in most cases. We first analyze the spike activities under the same inference time setting. For VGG-13, except the ReSU model with $T=5$ on CIFAR10, our SNN2ANN models perform $0.458\times$ $\sim$ $0.514\times$ activated spikes of the STBP-IF models and $0.588\times$ $\sim$ $0.64\times$ activated spikes of the STBP-PLIF models. For ResNet-17 with $T=5$, the spike activities of our SNN2ANN models are $0.412\times$ $\sim$ $0.553\times$of the STBP-IF,  and are $0.462\times$ $\sim$ $0.658\times$ of STBP-PLIF models.

\begin{figure*}[t]
\centering
\subfigure{
\includegraphics[width=0.3\textwidth]{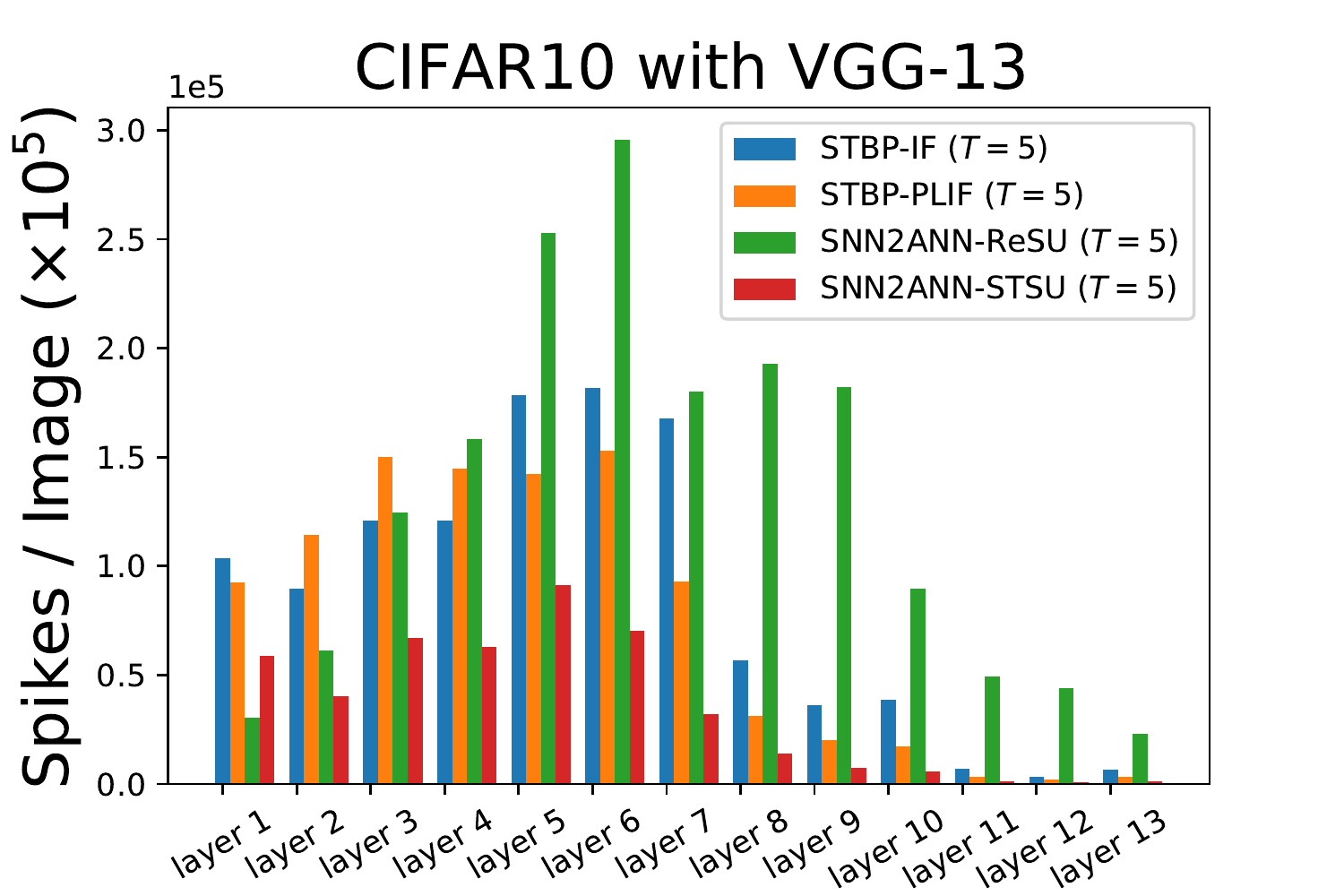}}
\subfigure{
\includegraphics[width=0.3\textwidth]{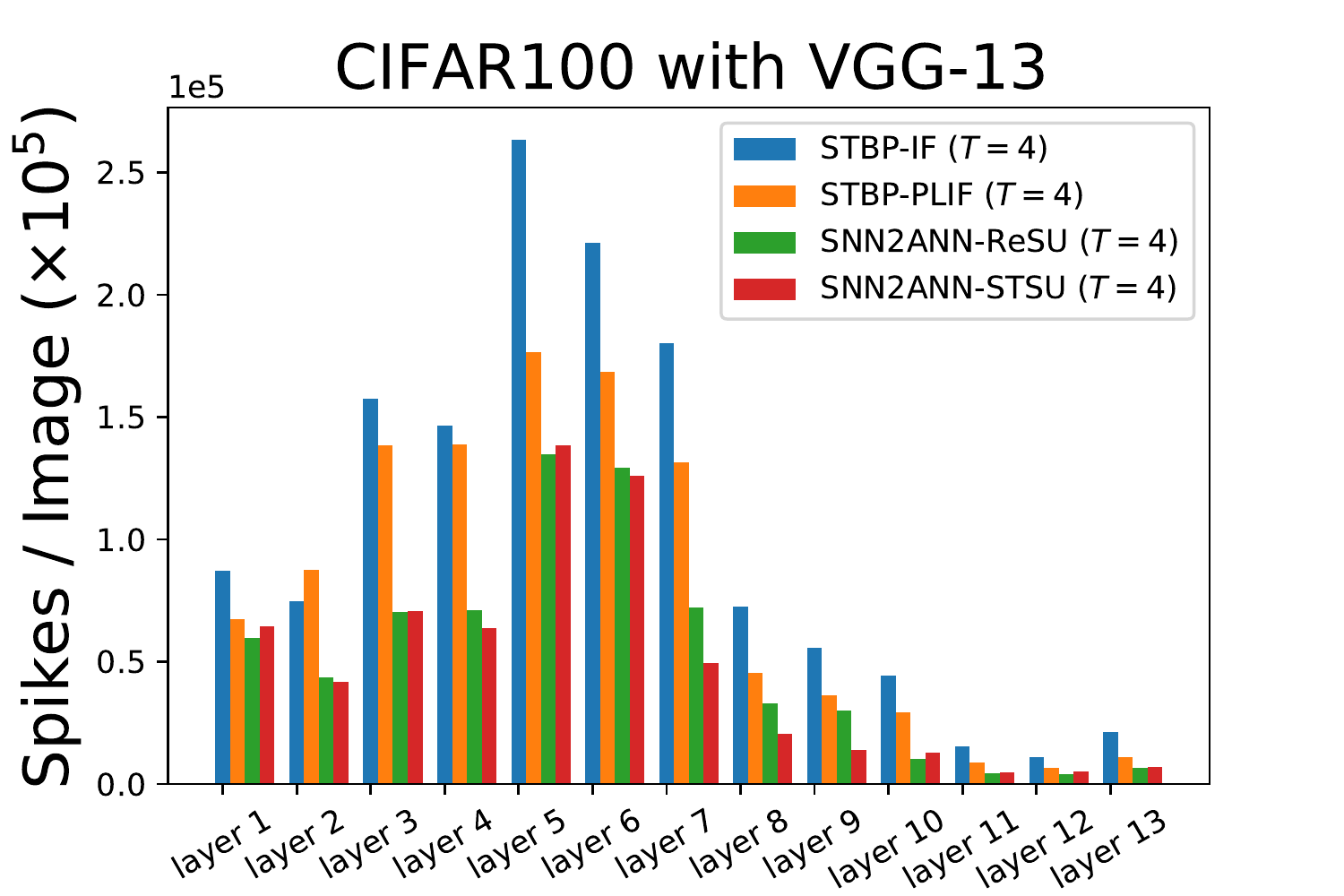}}
\subfigure{
\includegraphics[width=0.3\textwidth]{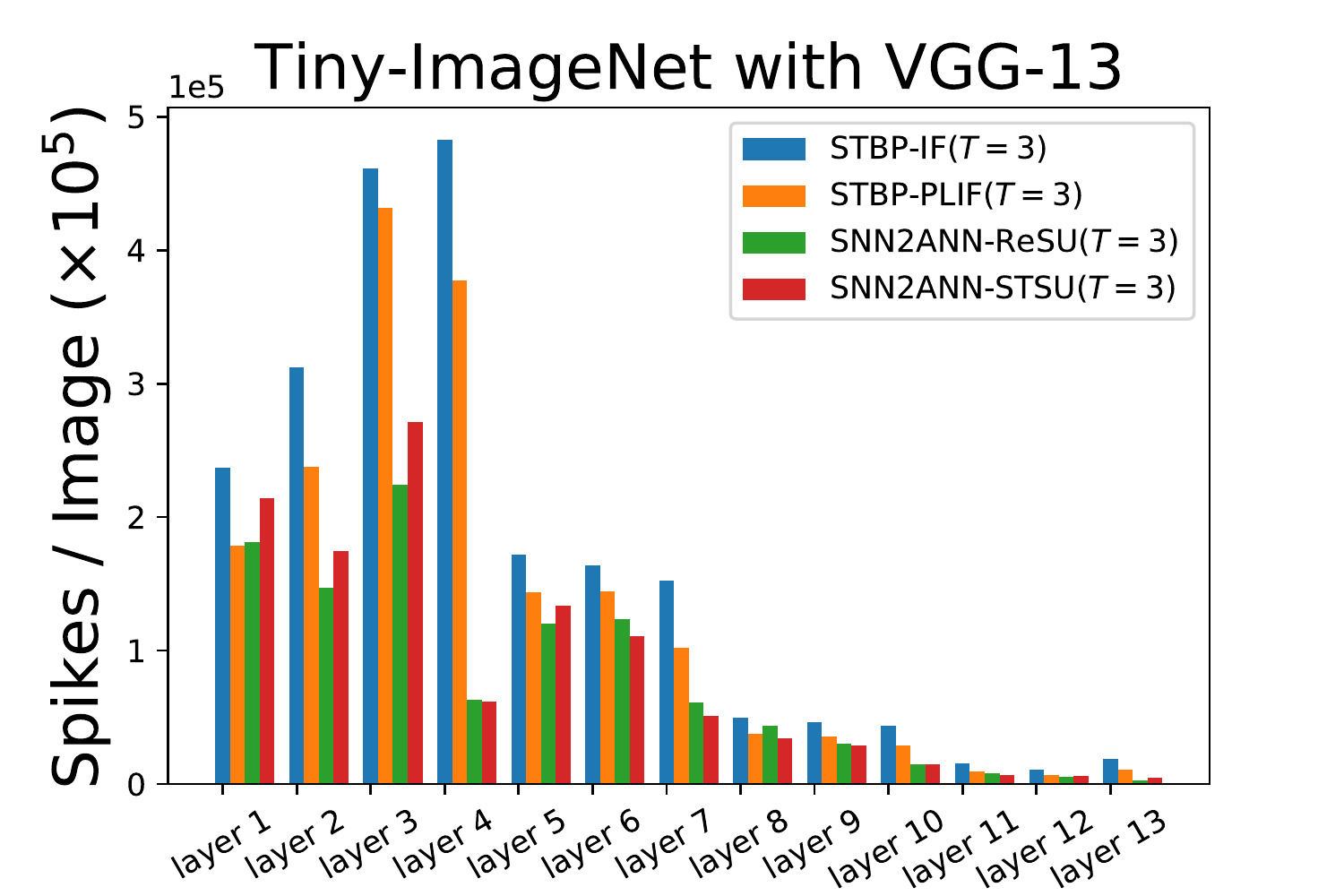}}
\subfigure{
\includegraphics[width=0.3\textwidth]{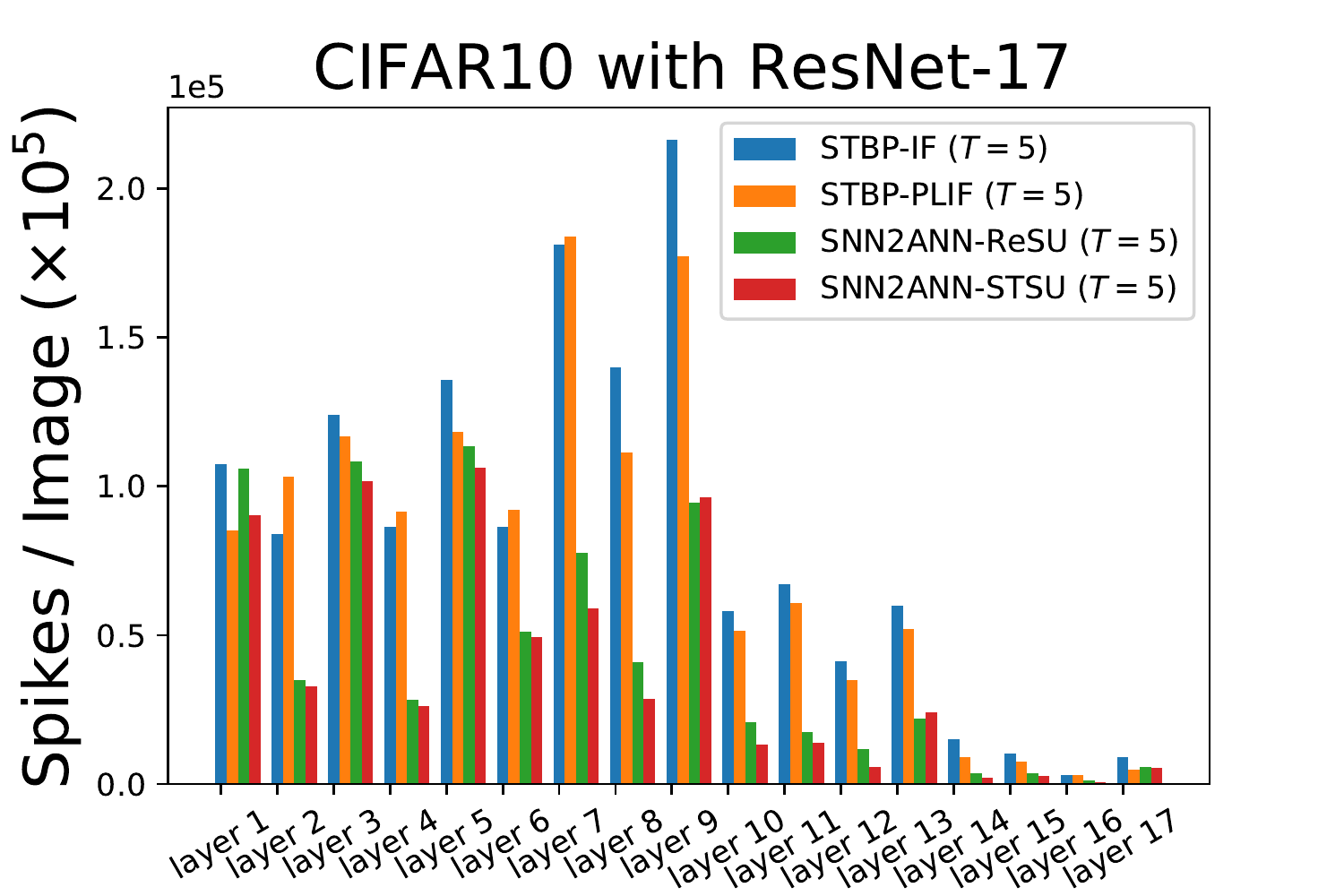}}
\subfigure{
\includegraphics[width=0.3\textwidth]{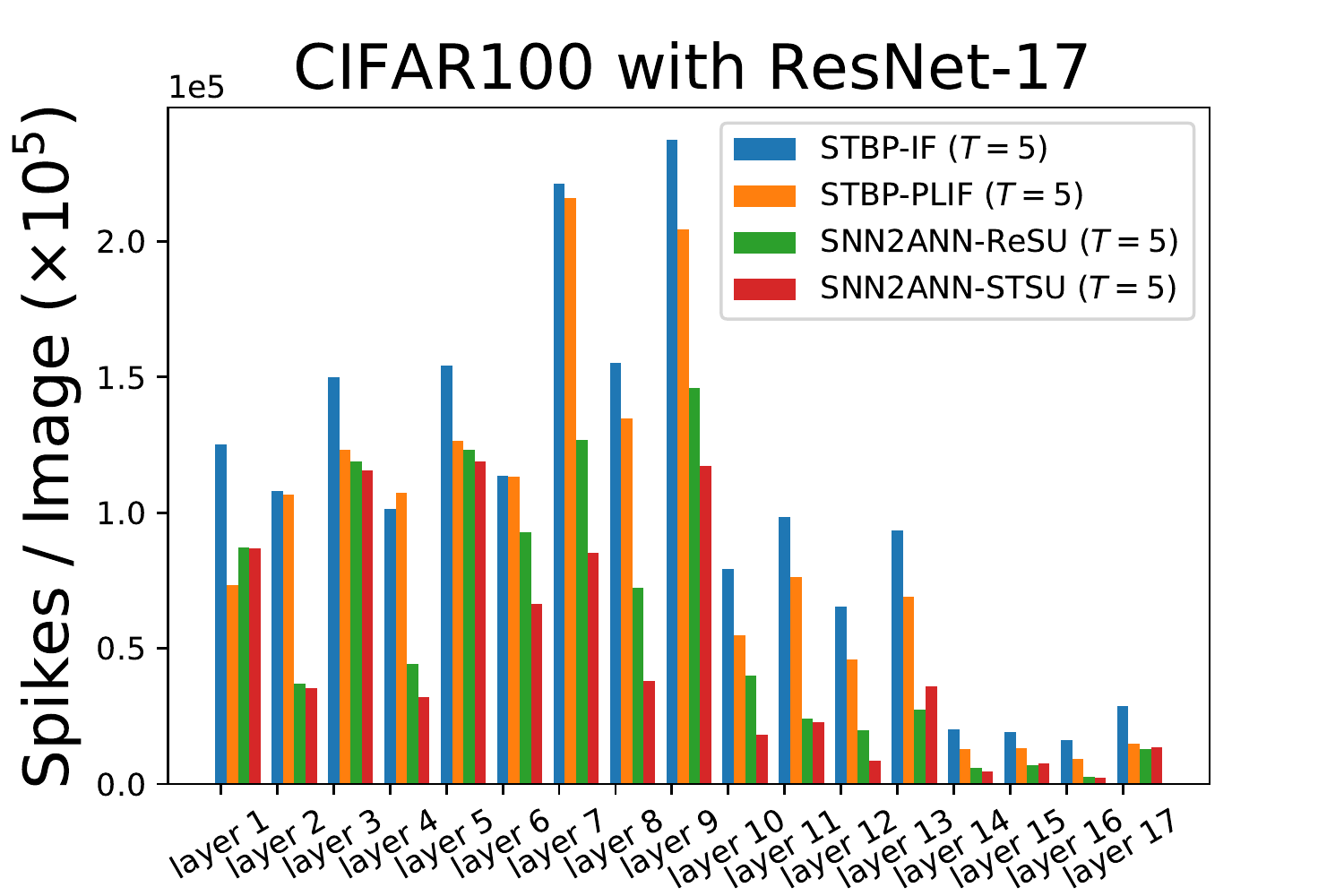}}
\subfigure{
\includegraphics[width=0.3\textwidth]{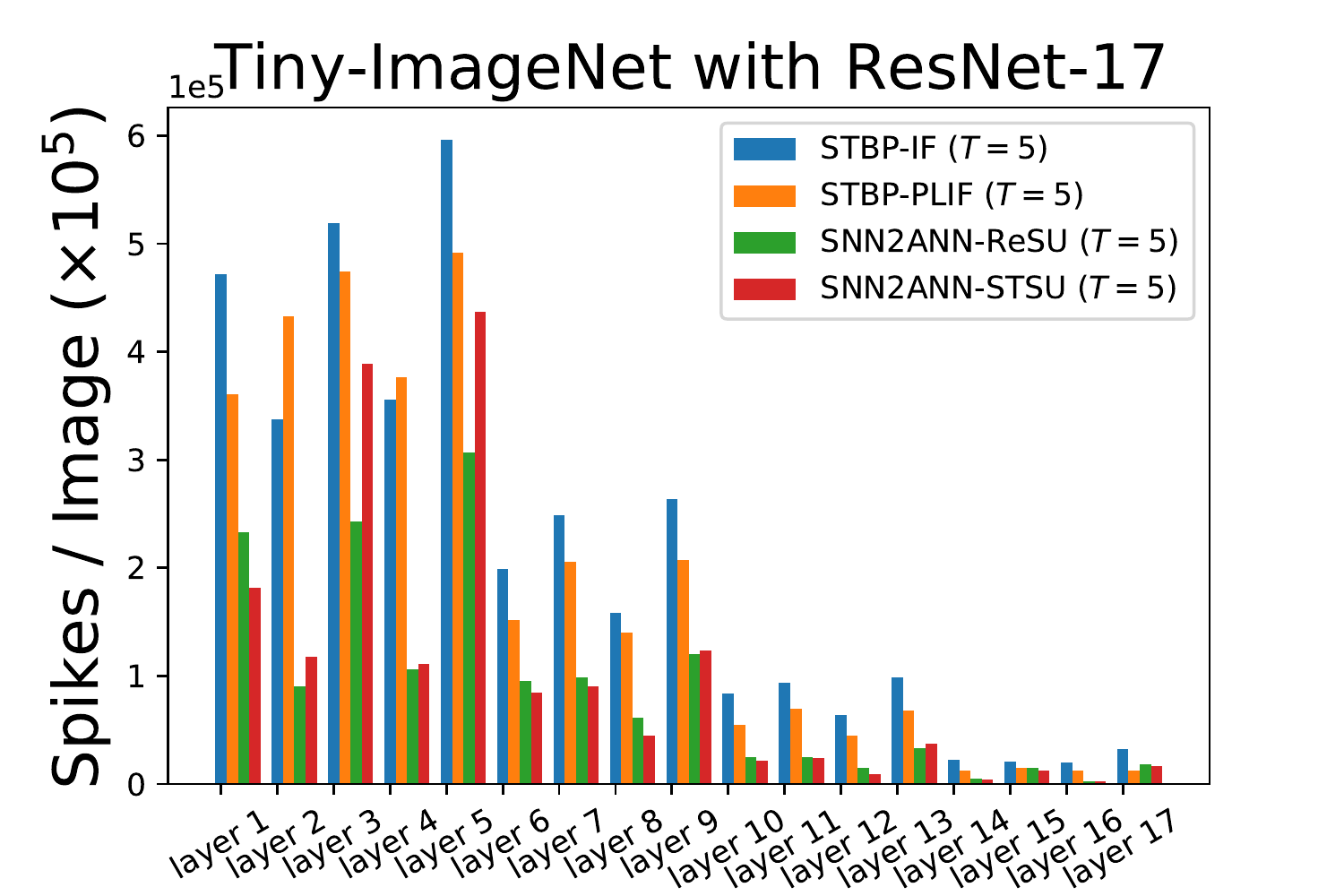}}
\caption{The layer-wise spike activities of SNN2ANN and STBP models. The abscissa is the layer number of the networks, and the ordinate is the activated spikes per image.}
\label{fig:spike_act}
\end{figure*}

Then, we compare the spike activities when SNN2ANN and STBP-based models have similar performance. Based on the \textbf{``Spikes/Image''} in Table~\ref{tab:same_cmp} of the STBP-PLIF and SNN2ANN models, for VGG-13, both SNN2ANN models achieve similar performance with STBP-PLIF under the same time steps setting, and the analysis has been mentioned above. For ResNet-17 on CIFAR10 and CIFAR100, the accuracy performance of the STBP-PLIF with $T=8$ is comparable with the SNN2ANN models. However, the spike activities of the SNN2ANN-based ResNet-17 are $0.282\times$ $\sim$ $0.407\times$ of the STBP-PLIF-based ResNet-17. Fig.~\ref{fig:spike_act} shows the layer-wise spike activities of SNN2ANN and STBP-based models on the benchmark datasets with the same time steps. The red and green bars are statistical spikes of SNN2ANN, while the bars with other colors denote the spikes of STBP-based models. The layer-wise spike activities of SNN2ANN models are less than STBP-based models in most cases.

\noindent\textbf{iii) Energy efficiency:}
Following \cite{DBLP:journals/corr/abs-2008-03658}, we estimate the compute energy of SNN by computing the energy benefit of SNN over ANN.
In the inference of SNNs, the floating-point (FP) MAC operations of ANNs are replaced by FP additions. In 45nm CMOS technology, the cost of the addition operation is $0.9pJ$, while the cost of MAC operation is $4.6pJ$ \cite{6757323,DBLP:journals/corr/abs-2008-03658}. The relationship between the number of operations per layer in SNN and ANN is
\begin{equation}
    \#\text{S}_{ops,n} = r_{n}\times \#\text{A}_{ops,n},
\end{equation}
where the spike rate $r_{n},n\in[1,L-1]$ denotes the number of spikes per neuron over all time steps in layer $n$. Layer $L$ is the classifier that satisfies $\#\text{S}_{ops,L} = \#\text{A}_{ops,L}$.
$\#\text{A}_{ops,n}$ is calculated based on
  \begin{equation}\label{ste:grad}
 \#\text{A}_{ops,n} =\left\{\begin{array}{l}
k_{w} \times k_{h} \times c_{i n} \times h_{o u t} \times w_{o u t} \times c_{o u t}, \text {Conv}, \\
 f_{i n} \times f_{\text {out }}, \text {FC}.
\end{array}\right.
\end{equation}
Let $E_{A}$ and $E_{S}$ represent the energy of ANN and SNN, respectively. We compute the energy benefits of SNN over ANN by

\begin{equation}
\small{
    \frac{E_{A}}{E_{S}}=\frac{4.6\times\sum\limits_{n=1}^{L}\#\text{A}_{ops,n}}{4.6\times(\#\text{En}_{ops}+\#\text{S}_{ops,L})+\sum\limits_{n=2}^{L-1}\#\text{S}_{ops,n}\times0.9},}
\end{equation}
where $\#\text{En}_{ops}=\#\text{S}_{ops,1}$ denotes the operations of the Encoder. The larger $\frac{E_{A}}{E_{S}}$ is, the more energy SNN saves.
The energy comparisons of ANN2SNN, STBP, and our SNN2ANN are given in Table~\ref{tab:energy}. The energies are estimated based on the best accuracy scores of RNL-RIL, STBP-PLIF, and SNN2ANN models. Since the SNN2ANN-based VGG-13 directly operates max-pooling on the convolution outputs, the spike rate is calculated based on the spiking feature maps of the pooling layer. Due to the RNL-RIL models requiring large time steps to achieve good performance, its energy efficiency cannot be comparable to STBP and SNN2ANN. Thanks to the low spike activities and small inference steps, our SNN2ANN models achieve considerable energy consumption in most cases. For VGG-13 on CIFAR10, the energy of ANN is $35.61\times$ of our SNN2ANN-STSU, outperforming the RNL-RIL and STBP-PLIF VGG-13. For ResNet-17 on all datasets, the SNN2ANN models have minimal energy overall comparisons.
\begin{table}[t]
  \centering
  \renewcommand{\arraystretch}{1.3}
  \caption{The compute energy of ANN (\textbf{A}) versus RNL-RIL (\textbf{R-R}), STBP-PLIF (\textbf{S-P}), and SNN2ANN (\textbf{S2A-ReSU/STSU}). }
  \scalebox{0.82}{
    \begin{tabular}{|c|c|c|c|c|c|}
    \hline
    \multirow{2}[4]{*}{\textbf{Dataset}} & \multirow{2}[4]{*}{\textbf{Network }} & \multicolumn{4}{c|}{\boldmath{}\textbf{Energy $\uparrow$}\unboldmath{}} \\
\cline{3-6}          &       & \boldmath{}\textbf{\textbf{$\text{A}/\text{R-R}$}}\unboldmath{} & \boldmath{}\textbf{\textbf{$\text{A}/\text{S-P}$ energy}}\unboldmath{} & \boldmath{}\textbf{\textbf{$\text{A}/\text{S2A-R}$ }}\unboldmath{} & \boldmath{}\textbf{\textbf{$\text{A}/\text{S2A-S}$ }}\unboldmath{} \\
    \hline
    \multirow{2}[4]{*}{\textbf{CIFAR10}} & VGG-13 & 2.93  & 26.50  & 2.68  & \textbf{35.61 } \\
\cline{2-6}          & ResNet-17  & -     & 6.37  & \textbf{19.97 } & \textbf{23.91 } \\
    \hline
    \multirow{2}[4]{*}{\textbf{CIFAR100}} & VGG-13 & 2.02  & \textbf{15.31 } & 12.70  & 13.62  \\
\cline{2-6}          & ResNet-17 & -     & 4.33  & \textbf{12.48 } & \textbf{14.65 } \\
    \hline
    \multirow{2}[4]{*}{\textbf{Tiny-ImageNet}} & VGG-13 & 1.23  & \textbf{15.29 } & 13.57  & 13.20  \\
\cline{2-6}          & ResNet-17  & -     & 5.98  & \textbf{11.07 } & \textbf{11.42 } \\
    \hline
    \end{tabular}%
    }
  \label{tab:energy}%
\end{table}%
\subsection{Ablation Study}
Table~\ref{tab:ab_study} presents the ablation study of the SNN2ANN. The accuracies are reported from the best performance. Both STSU and ReSU promote the SNNs for classification. For VGG-13, the ReSU-based SNNs achieve $89.17\%$, $ 57.13\%$, $50.69\%$ on CIFAR10, CIFAR100, and Tiny-ImageNet, respectively, while the STSU-based SNNs perform $89.86\%$, $42.50\%$, and $50.79\%$ on those datasets. For ResNet-17, the accuracy of CIFAR10, CIFAR100, and Tiny-ImageNet are $89.30\%$, $68.14\%$, and $51.82\%$ for the ReSU-based SNNs, are $89.46\%$, $68.06\%$, and $50.15\%$ for the STSU-based SNNs. By introducing the ATA to adjust the firing thresholds, the accuracies of all models are enhanced significantly.
\begin{table}[t]
  \centering
  \renewcommand{\arraystretch}{1.1}
  \caption{Ablation study of SNN2ANN models.}
  \scalebox{0.75}{
    \begin{tabular}{|c|c|c|c|c|c|c|c|c|}
    \hline
          & \multicolumn{4}{c|}{\textbf{VGG-13}} & \multicolumn{4}{c|}{\textbf{ResNet-17}} \\
    \hline
    \textbf{Dataset} & \textbf{STSU} & \textbf{ReSU} & \textbf{ATA} & \textbf{Acc (\%)} & \textbf{STSU} & \textbf{ReSU} & \textbf{ATA} & \textbf{Acc (\%)} \\
    \hline
    \multirow{4}{*}{\textbf{CIFAR10}}
    &  &       &       & 10.41 &  &       &       & 11.33 \\\cline{2-9}
    & \checkmark &       &       & 89.86 & \checkmark &       &       & 89.46 \\
\cline{2-9}          &       & \checkmark &       & 89.17 &       & \checkmark &       & 89.30 \\
\cline{2-9}          & \checkmark &       & \checkmark & \textbf{92.29} & \checkmark &       & \checkmark & \textbf{92.79} \\
\cline{2-9}          &       & \checkmark & \checkmark & \textbf{92.67} &       & \checkmark & \checkmark & \textbf{92.88} \\
    \hline
    \multirow{4}{*}{\textbf{CIFAR100}}
    &  &       &       & 1.17 &  &       &       & 1.08 \\
\cline{2-9}
    & \checkmark &       &       & 42.50 & \checkmark &       &       & 68.06 \\
\cline{2-9}          &       & \checkmark &       & 57.13 &       & \checkmark &       & 68.14 \\
\cline{2-9}          & \checkmark &       & \checkmark & \textbf{69.31} & \checkmark &       & \checkmark & \textbf{73.61} \\
\cline{2-9}          &       & \checkmark & \checkmark & \textbf{71.19} &       & \checkmark & \checkmark & \textbf{73.14} \\
    \hline
    \multirow{4}{*}{\textbf{Tiny-ImageNet}}
    &  &       &       & 0.43 &  &       &       & 0.60 \\
\cline{2-9}
    & \checkmark &       &       & 50.79 & \checkmark &       &       & 50.15 \\
\cline{2-9}          &       & \checkmark &       & 50.69 &       & \checkmark &       & 51.82 \\
\cline{2-9}          & \checkmark &       & \checkmark & \textbf{54.39} & \checkmark &       & \checkmark & \textbf{57.02} \\
\cline{2-9}          &       & \checkmark & \checkmark & \textbf{55.27} &       & \checkmark & \checkmark & \textbf{56.31} \\
    \hline
    \end{tabular}%
    }
  \label{tab:ab_study}%
\end{table}%
 \begin{figure}[t]
    \centering
    \includegraphics[width=0.5\textwidth]{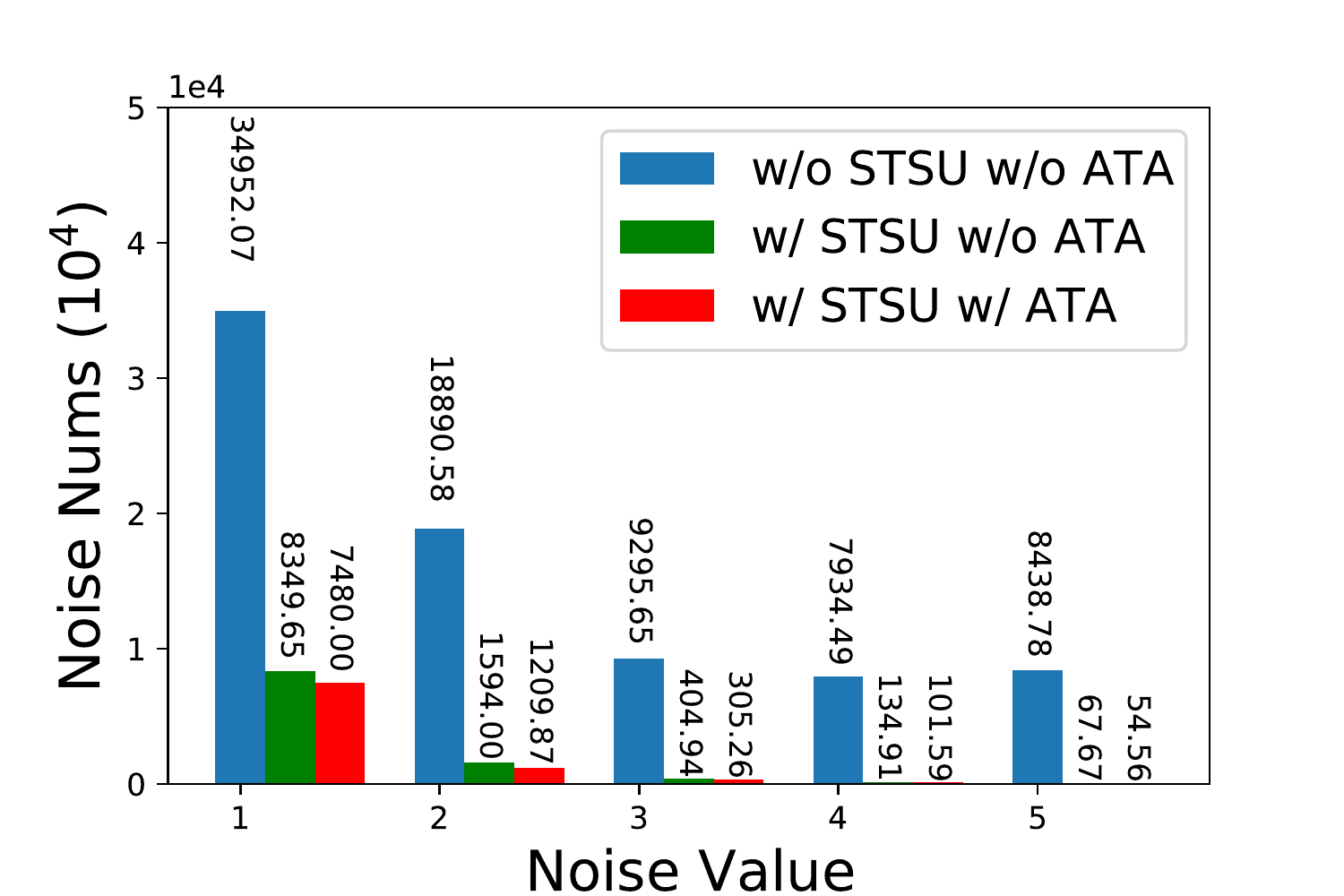}
    \caption{The number of noisy spikes comparison between w/ STSU w/ ATA, w/ STSU w/o ATA, and w/o STSU w/o ATA.}
    \label{fig:noise_cmp}
\end{figure}

To demonstrate that the ATA decreases the firing of noise spikes, we compare the layer-wise noisy spikes number of the w/ STSU w/ ATA with w/ STSU w/o ATA and w/o STSU w/o ATA on the CIFAR100 dataset. Experiment results are presented in Fig. \ref{fig:noise_cmp}. The max value of the noisy spikes is the size of the time window, which is $5$ in this instance. The ``w/o STSU w/o ATA'' generates the most noises, and the ANN2SNN fails to train the model. Introducing the STSU for the ANN branch modeling, ``w/ STSU w/o ATA'' reduces the noisy spikes, especially for the noises which have large values. Then, the accuracy is enhanced to $68.06\%$. Finally, we combine the STSU with ATA for SNN2ANN training, and the ``w/ STSU w/ ATA'' further decreases the noisy spikes, increasing the accuracy to $73.61\%$.
\subsection{Equivalent Between ANN and SNN in SNN2ANN}
\begin{figure*}
\centering
\subfigure{
\includegraphics[width=0.3\textwidth]{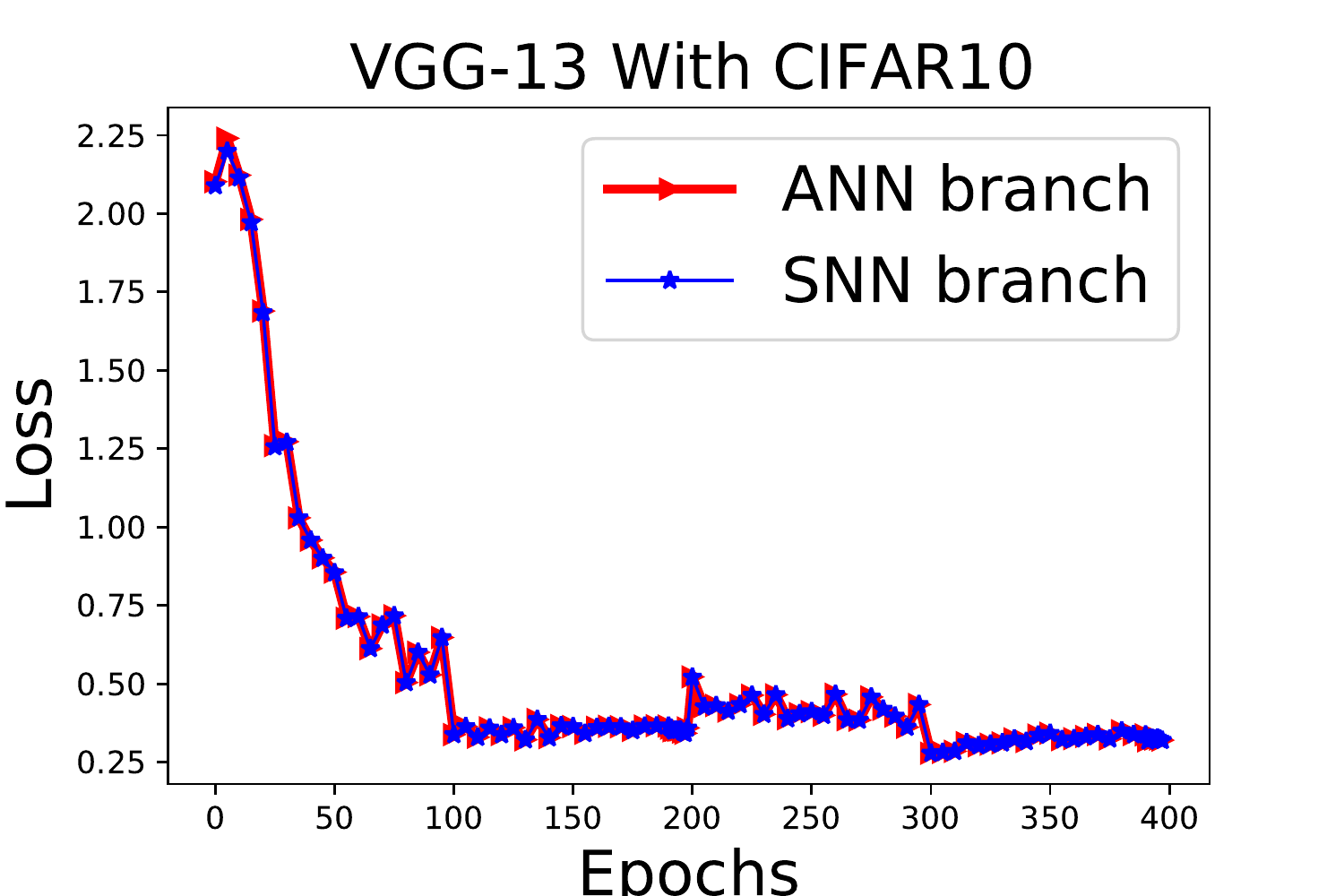}}
\subfigure{
\includegraphics[width=0.3\textwidth]{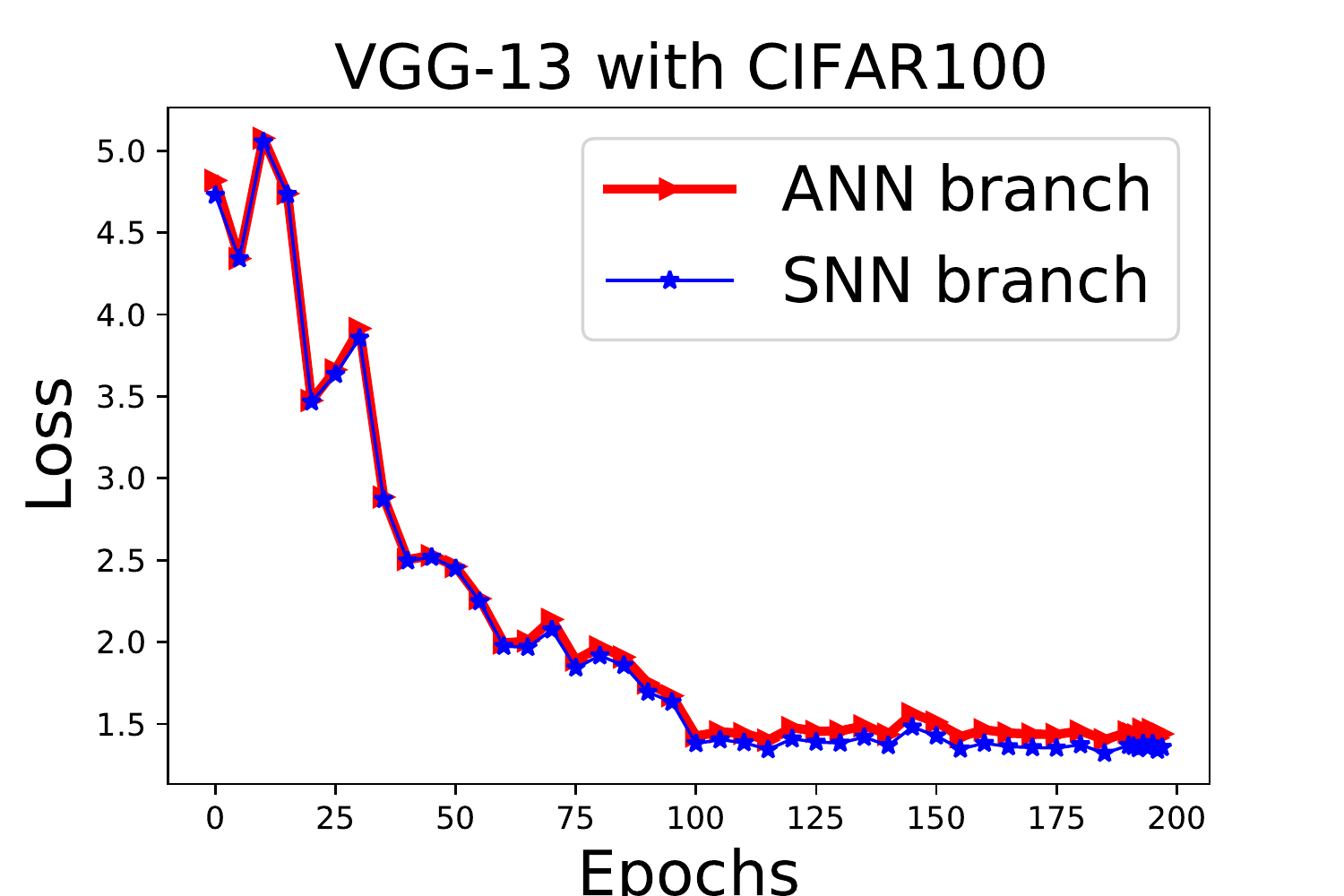}}
\subfigure{
\includegraphics[width=0.3\textwidth]{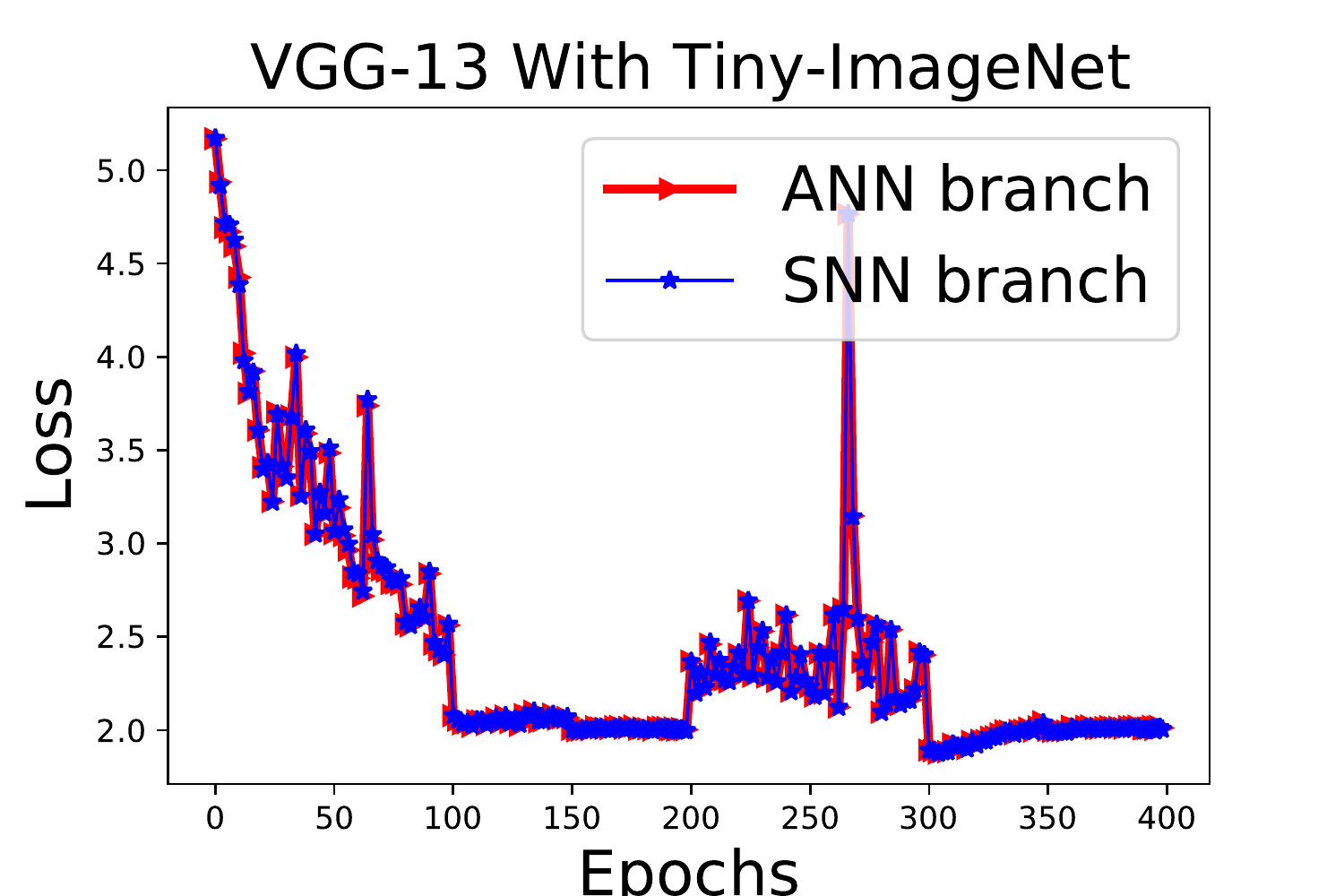}}
\subfigure{
\includegraphics[width=0.3\textwidth]{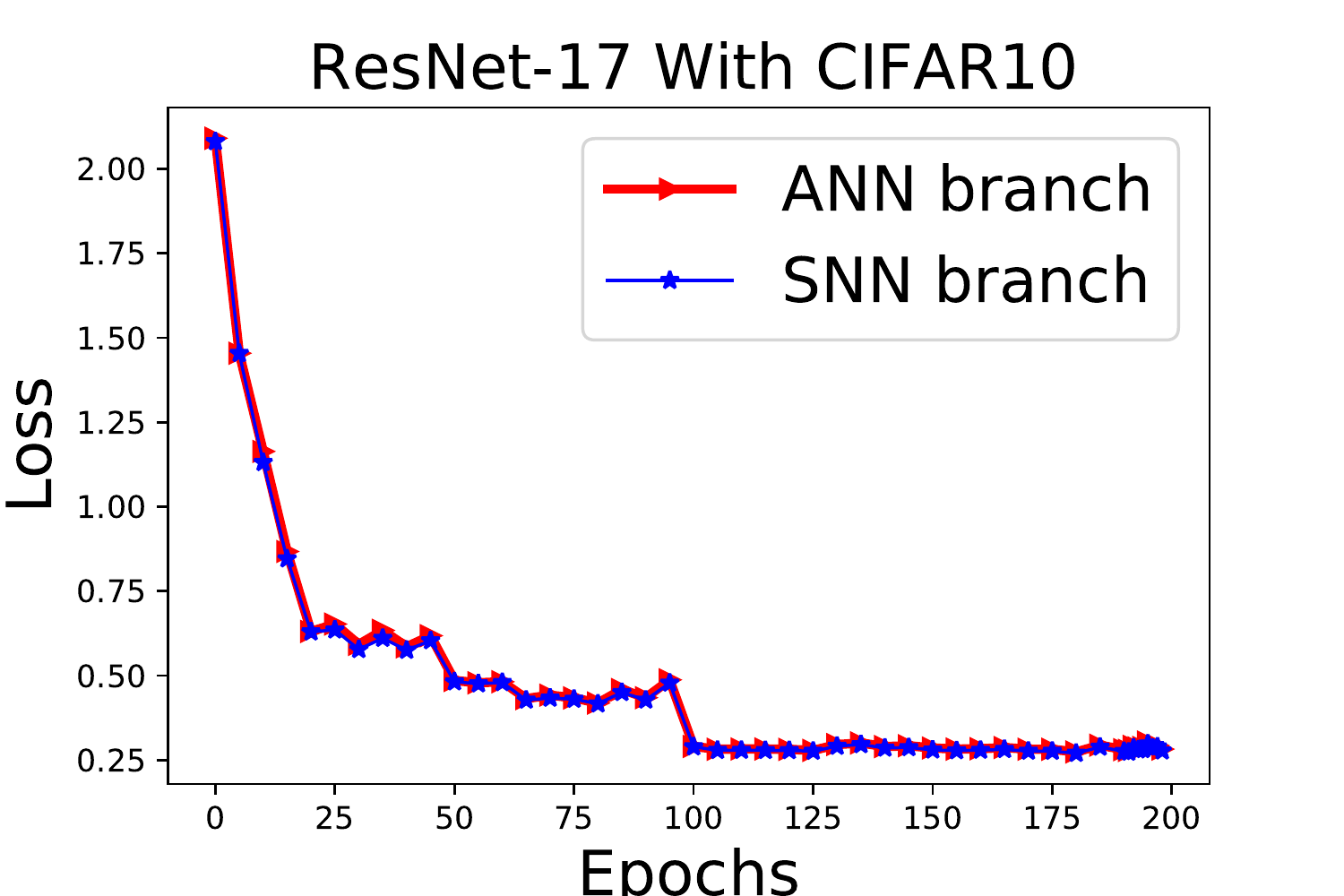}}
\subfigure{
\includegraphics[width=0.3\textwidth]{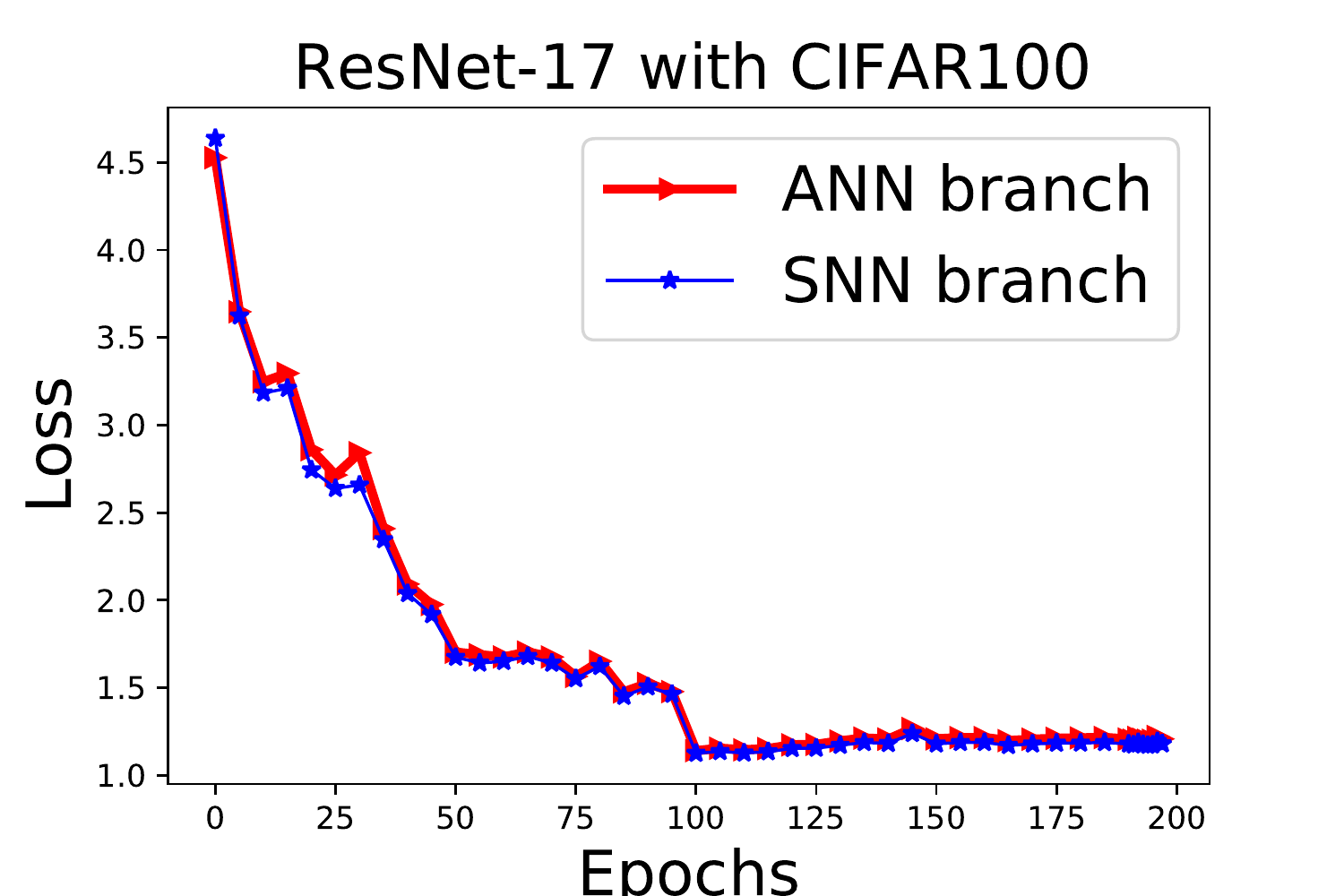}}
\subfigure{
\includegraphics[width=0.3\textwidth]{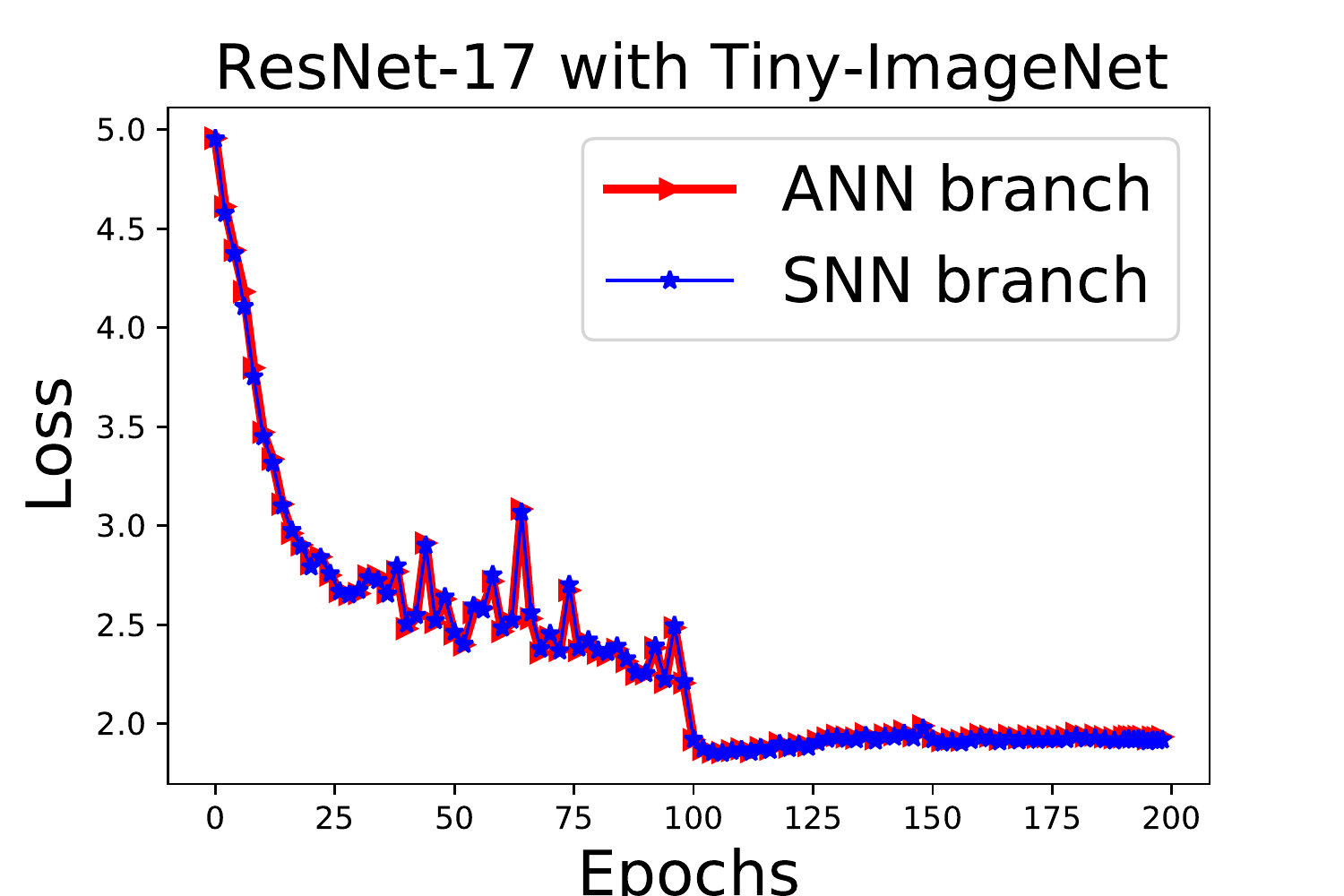}}
\caption{The validation loss of the ReSU-based ANN and SNN branches on the benchmark datasets. The abscissa is the layer number of the networks, and the ordinate denotes the loss value.}
\label{fig:val_loss}
\end{figure*}
\begin{figure*}
\centering
\subfigure{
\includegraphics[width=0.3\textwidth]{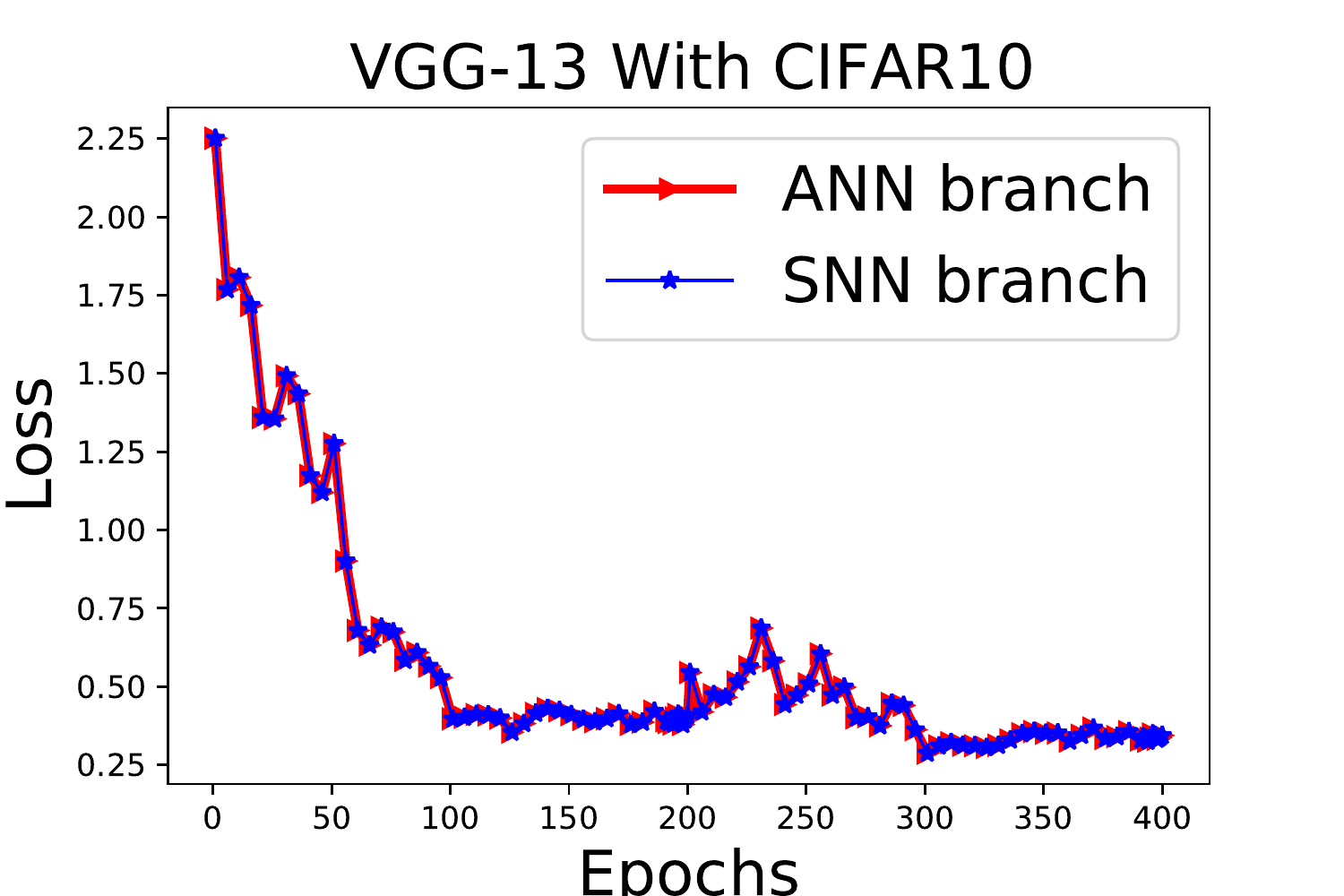}}
\subfigure{
\includegraphics[width=0.3\textwidth]{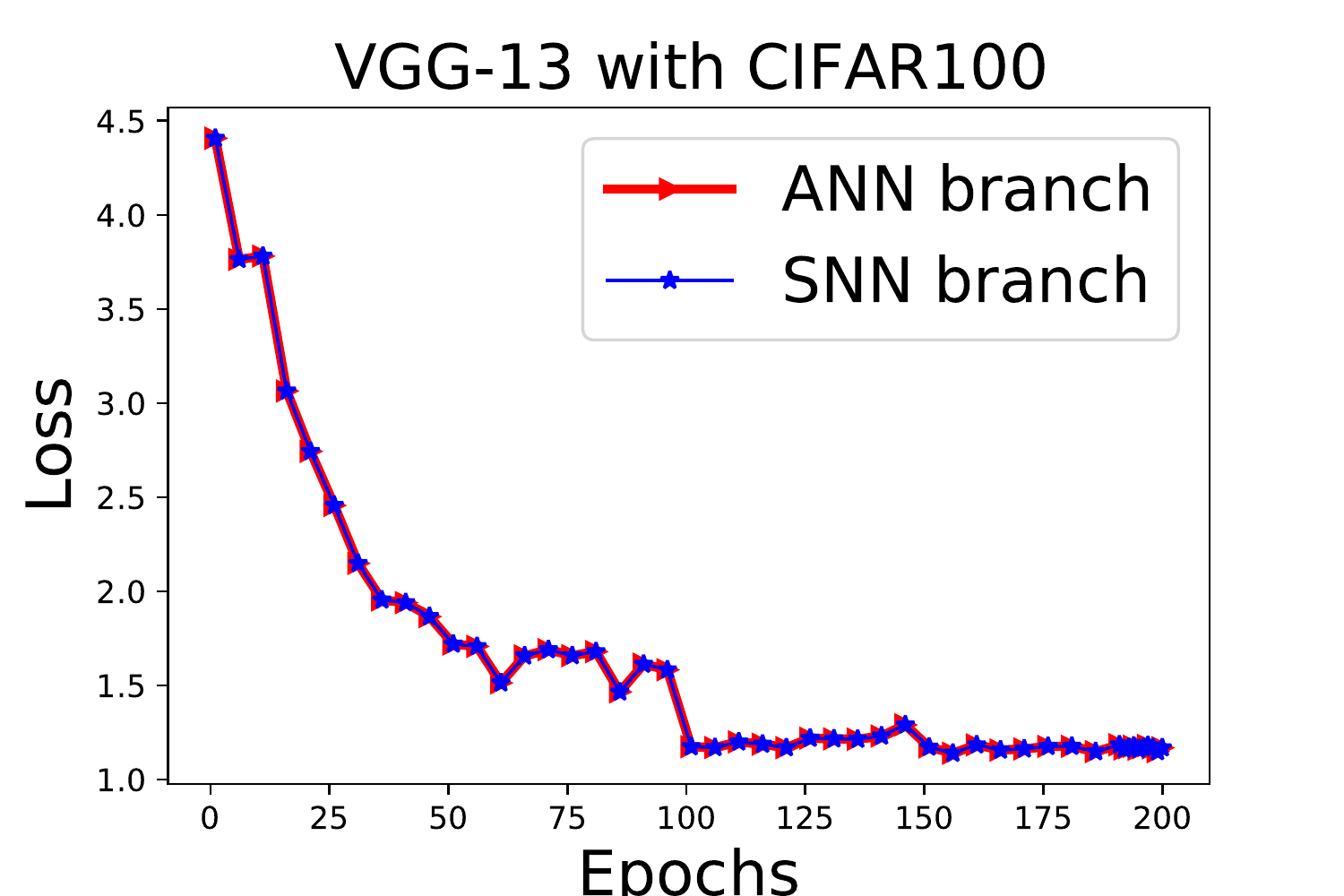}}
\subfigure{
\includegraphics[width=0.3\textwidth]{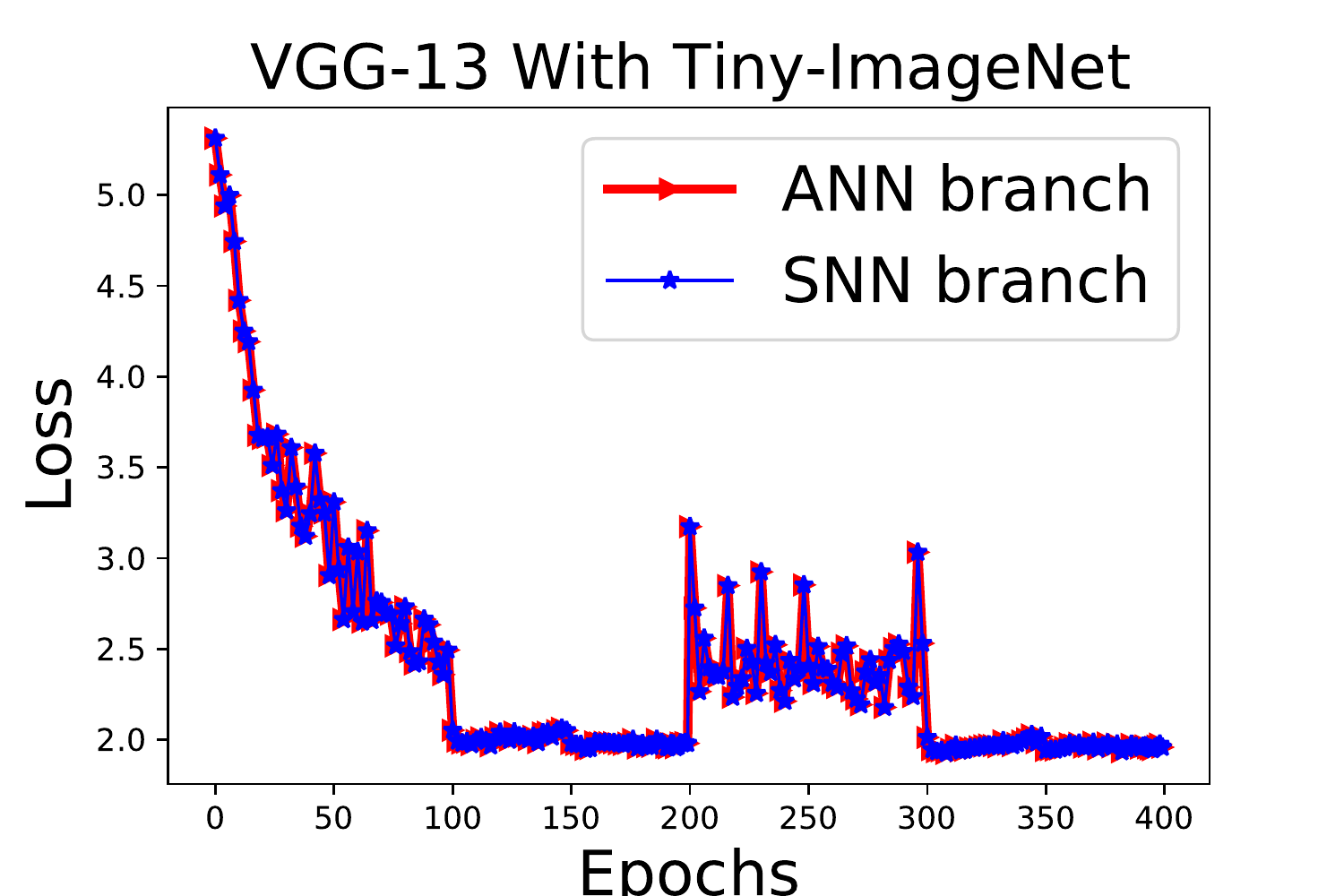}}
\subfigure{
\includegraphics[width=0.3\textwidth]{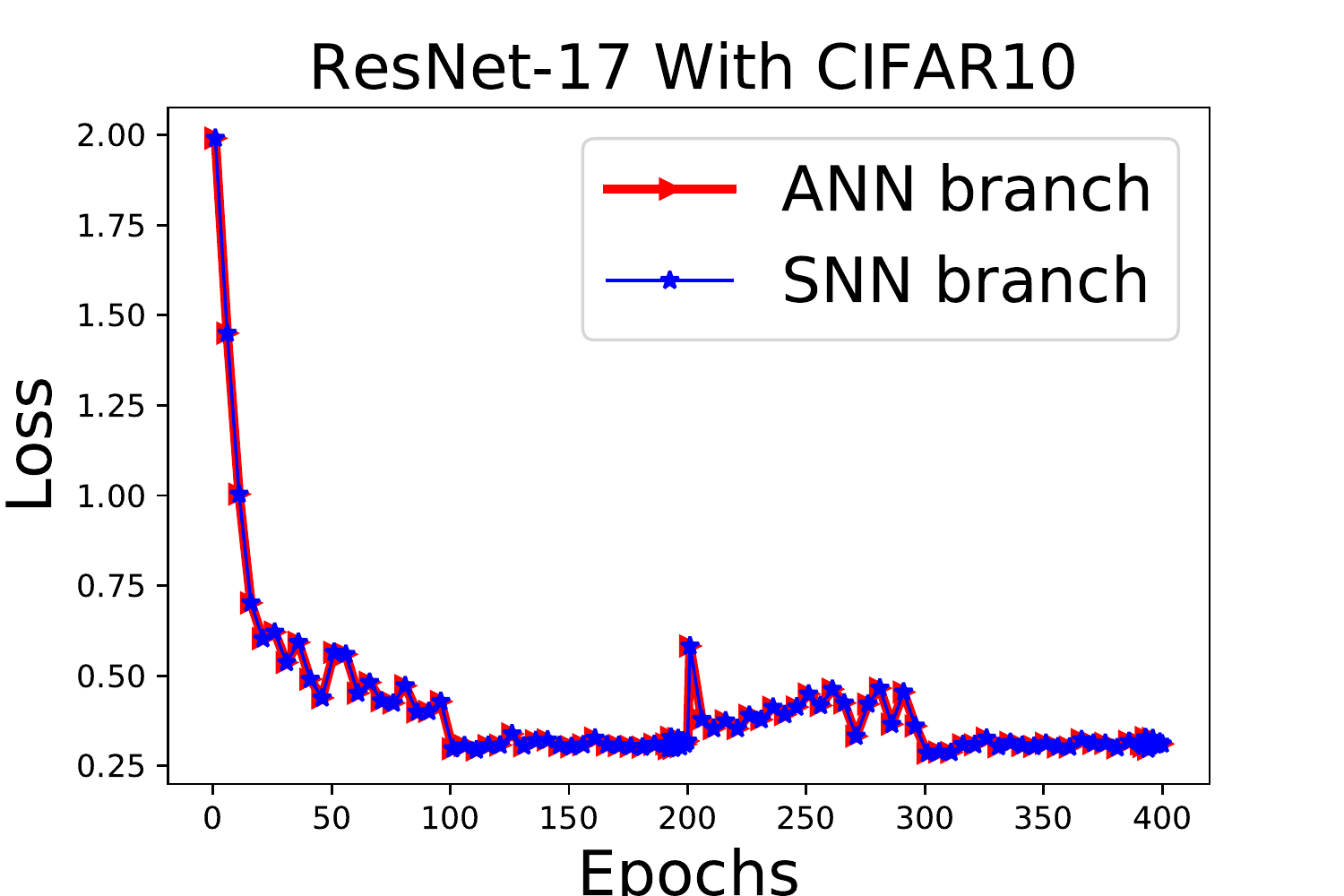}}
\subfigure{
\includegraphics[width=0.3\textwidth]{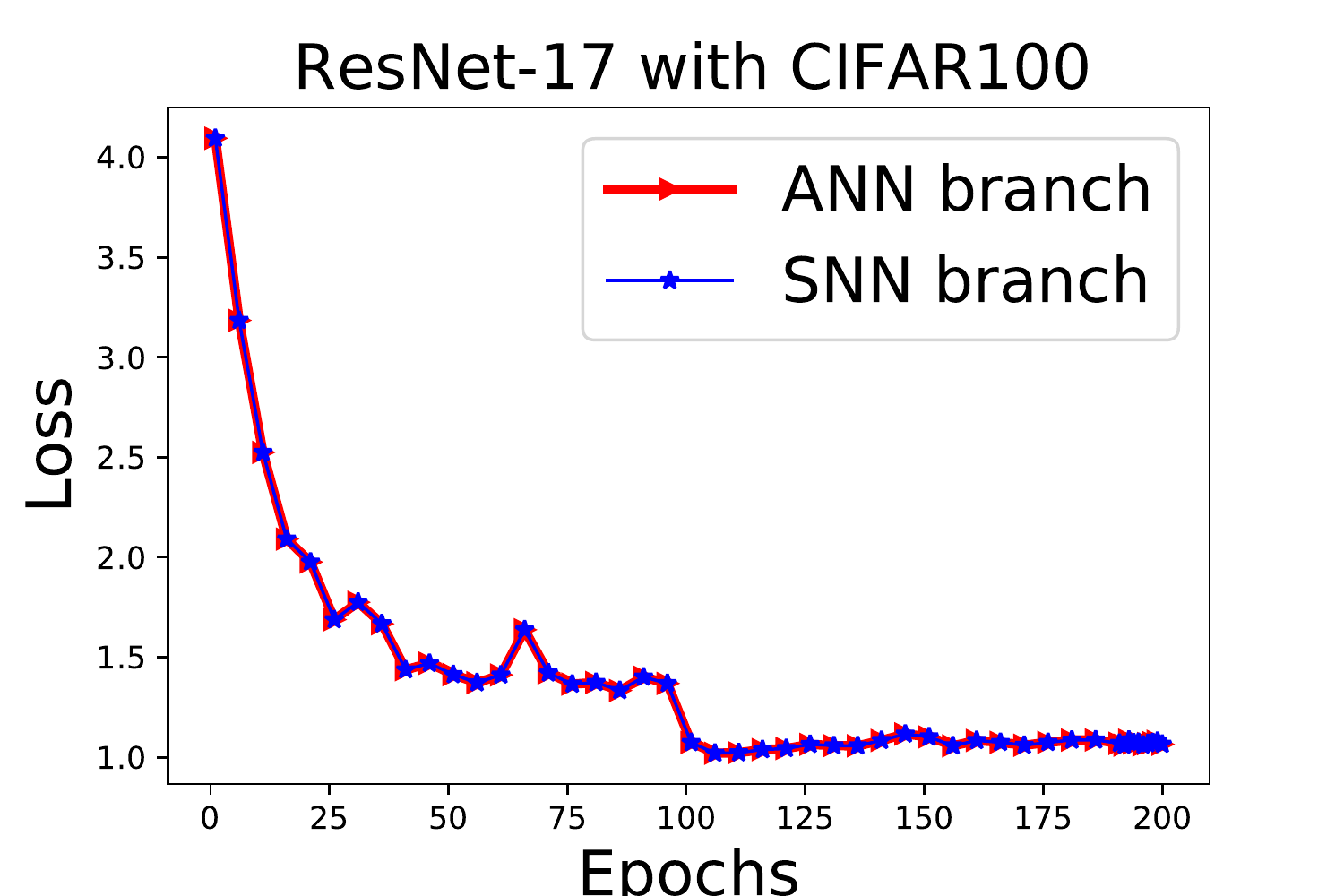}}
\subfigure{
\includegraphics[width=0.3\textwidth]{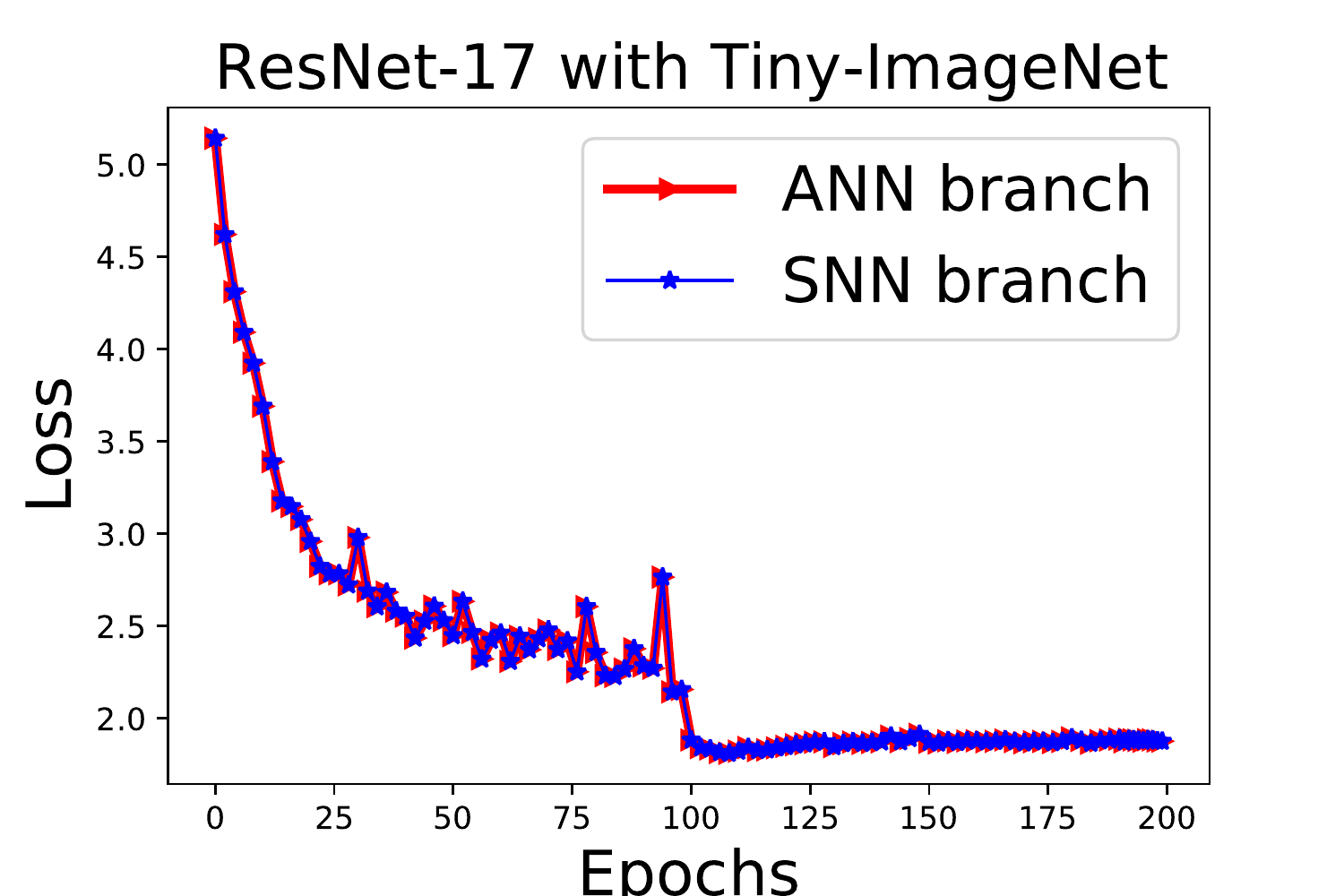}}
\caption{The validation loss of the STSU-based ANN and SNN branches on the benchmark datasets. The abscissa is the layer number of the networks, and the ordinate denotes the loss value.}
\label{fig:val_loss2}
\end{figure*}
To show the equivalent between ANN and SNN branches, we visualize the validation loss of both branches through the training process. Fig.~\ref{fig:val_loss} and~\ref{fig:val_loss2} display show that SNN loss changing curves are consistent with the ReSU-based and STSU-based ANNs. It demonstrates that the ReSU and STSU map the accumulated spikes as the activation values of the ANN branch, making the ReSU/STSU-based ANN can represent the SNN. Though the BP training is only worked on the ANN branch, the decrease of the validation curves of SNN branches indicates that the SNN2ANN success in transferring the training of SNN on the ReSU/STSU-based ANN. In addition, the accuracy changing curve of SNN2ANN in Fig.~\ref{fig:curve_cmp} also validates the convergence and effectiveness of the SNN2ANN training.
\vspace{-0.1cm}
\section{Conclusion}
This paper proposes the SNN2ANN framework to train the SNN in a fast and memory-efficient way. The SNN2ANN enables the BP algorithm on the ANN branch, and the weight-sharing mechanism guarantees that both ANN and SNN branches are updated simultaneously. Since the ReSU/STSU maps the spiking features of the SNN on the ANN branch, the classification error of the SNN can be optimized by training the ANN branch. Moreover, the adaptive threshold adjustment addresses the noisy spike problem and improves the performance of SNNs. Experiment results demonstrate that our SNN2ANN models achieve considerable accuracy with fast training, low memory cost, sparse spike activities, and fast inference.
\bibliographystyle{IEEEtran}
\bibliography{manuscript}
\end{document}